\documentclass[lettersize,journal]{IEEEtran}
\usepackage{amsmath, amssymb, amsfonts, wasysym}
\usepackage{graphicx, wrapfig}
\usepackage{float}
\usepackage[ruled,vlined,linesnumbered]{algorithm2e}
\usepackage{booktabs}
\usepackage[font=small]{caption}
\usepackage{xcolor,colortbl}
\usepackage[colorlinks=true, pdfborder={0 0 0}, linkcolor=magenta]{hyperref}
\let\labelindent\relax
\usepackage{enumerate}
\usepackage{enumitem}

\usepackage{amsthm}
\usepackage{multirow}

\newcommand{\bb}[1]{\boldsymbol{#1}}
\SetKw{And}{\textbf{and}}
\SetKw{Or}{\textbf{or}}
\SetKw{Not}{\textbf{not}}
\SetKw{Is}{\textbf{is}}
\SetKw{True}{\textbf{true}}
\SetKw{False}{\textbf{false}}
\SetKw{Return}{\textbf{return}}
\SetKw{Reached}{\textbf{reached}}
\SetKw{Advanced}{\textbf{advanced}}
\SetKw{Trapped}{\textbf{trapped}}
\SetKw{Continue}{\textbf{continue}}
\SetKw{Break}{\textbf{break}}
\SetKwRepeat{Do}{do}{while}

\usepackage{etoolbox}
\makeatletter
\patchcmd{\@algocf@start}
{-1.5em}
{0pt}
{}{}
\makeatother

\newtheorem{theorem}{Theorem}

\newtheorem{problem}{Problem}[section]
\newtheorem{corollary}{Corollary}
\newtheorem{definition}{Definition}[section]
\theoremstyle{definition}

\title{\LARGE \bf
	Real-Time Sampling-Based Safe Motion Planning for Robotic Manipulators in Dynamic Environments
}

\author{Nermin Covic$^{1}$, Bakir Lacevic$^{1}$, Dinko Osmankovic$^{2}$ and Tarik Uzunovic$^{1}$
	\thanks{$^{1}$Authors are with the Faculty of Electrical Engineering, University of Sarajevo, Bosnia and Herzegovina, {\tt\small \{nermin.covic, bakir.lacevic, tarik.uzunovic\}@etf.unsa.ba}.}%
	\thanks{$^{2}$Author is with Agile Robots SE, Munich, Germany, {\tt\small dinko.osmankovic@agile-robots.com}.}%
}

\graphicspath{{Figures/}} 

\begin{document}
\maketitle
\thispagestyle{empty}
\pagestyle{empty}


\begin{abstract}
	In this paper, we present the main features of Dynamic Rapidly-exploring Generalized Bur Tree (DRGBT) algorithm, a sampling-based planner for dynamic environments. We provide a detailed time analysis and appropriate scheduling to facilitate a real-time operation. To this end, an extensive analysis is conducted to identify the time-critical routines and their dependence on the number of obstacles. Furthermore, information about the distance to obstacles is used to compute a structure called \textit{dynamic expanded bubble} of free configuration space, which is then utilized to establish sufficient conditions for a guaranteed safe motion of the robot while satisfying all kinematic constraints. An extensive comparative study is conducted to compare the proposed algorithm to competing state-of-the-art methods. Finally, an experimental study on a real robot is carried out covering a variety of scenarios including those with human presence. The results show the effectiveness and feasibility of real-time execution of the proposed motion planning algorithm within a typical sensor-based arrangement, using cheap hardware and sequential architecture, without the necessity for GPUs or heavy parallelization. 
\end{abstract}


\vspace{-0.4cm}
\section{Introduction}
\vspace{-0.05cm}
Achieving real-time motion planning (RTMP) for robotic manipulators is challenged by highly dynamic, uncertain, and unpredictably changing environments. Robotic manipulators usually have six or more degrees of freedom (DoF), which makes such planning in relatively high dimensional configuration space ($\mathcal{C}$-space) difficult, even for seemingly simple scenarios. Since obstacle positions and their motions are not fully known a priori, it is desirable that the robot reacts in real time with respect to captured changes in the environment. The planned trajectories, or their fragments, should accordingly be checked for validity and modified when necessary. 
Since any change in the environment must be detected as quickly as possible, and the robot must avoid any undesired contact with incoming obstacles, each plan needs to be computed relatively fast enough while accounting for inherent constraints, e.g., maximal velocity, acceleration, etc. A broad spectrum of methods have been proposed for the last four decades that aim at dealing with dynamic environments (DEs). 

One of the first approaches to tackle DEs is using artificial potential fields (APFs) (e.g., \cite{khatib1986real}) since they enable fast computation of plans.
However, most of APF-based methods suffer from local-minima problem. This triggered the development of alternative global-oriented algorithms, of which sampling-based (SB) methods gained considerable popularity.

Incorporation of SB methods into DEs is firstly attempted through \cite{leven2001toward} and \cite{leven2002framework}. These approaches are closely related to probabilistic roadmap methods (PRMs) \cite{kavraki1996probabilistic}, differing in a preprocessing stage that can be easily modified to account for changes in the environment. This modification is completed in real time by mapping cells from the workspace to nodes and arcs in a graph. A similar approach is used in \cite{jaillet2004prm}, where \textit{lazy-evaluation} mechanisms are used to update the graph. Moreover, a single-query method (e.g., \cite{lavalle1998rapidly}) may be used as a local planner in order to update the roadmap rapidly.

Neural networks find their place within MP, where \cite{yang2000efficient} proposes an RTMP method aiming at safety considerations in DEs. The dynamic neural activity landscape of the biologically inspired neural network is used to construct and devise further plans. Decomposition-based MP is proposed in \cite{brock2001decomposition}, where the original planning problem is decomposed into simpler subproblems. An experimental study with an 11-DoF mobile manipulator proves the real-time operation of this method.

\begin{table*}[t]
\centering
\caption{Summary of the relevant features of the selected set of the state-of-the-art algorithms.}
\vspace{-0.15cm}
\label{tab_features_comparison}
\resizebox{\textwidth}{!}{%
	\begin{tabular}{@{}lllllllllllll@{}}
		\toprule
		\begin{tabular}[c]{@{}l@{}}
			Algorithm
		\end{tabular} &
		\begin{tabular}[c]{@{}l@{}}
			Replanning\\ 
			technique\\
			(based on)
		\end{tabular} &
		\begin{tabular}[c]{@{}l@{}}
			Real-time\\ 
			(schedulabi-\\ 
			lity analysis)
		\end{tabular} &
		\begin{tabular}[c]{@{}l@{}}
			Max. tested\\ 
			replanning \\
			freq. $\mathrm{[Hz]}$
		\end{tabular} &
		\begin{tabular}[c]{@{}l@{}}
			Trajectory\\ 
			generation/\\
			interpolation
		\end{tabular} &
		\begin{tabular}[c]{@{}l@{}}
			Imposed\\ 
			constraints
		\end{tabular} &
		\begin{tabular}[c]{@{}l@{}}
			Guaran-\\ 
			teed safe\\ 
			motion
		\end{tabular} &
		\begin{tabular}[c]{@{}l@{}}
			Max. obs.\\
			velocity\\
			$\mathrm{[m/s]}$
		\end{tabular} &
		\begin{tabular}[c]{@{}l@{}}
			Validation in\\ 
			simulation
		\end{tabular} &
		\begin{tabular}[c]{@{}l@{}}
			Experimental\\ 
			validation
		\end{tabular} &
		\begin{tabular}[c]{@{}l@{}}
			Perception\\
			(if sensor-\\
			based)
		\end{tabular} &
		\begin{tabular}[c]{@{}l@{}}
			Hardware\\
			(implemen-\\
			tation)
		\end{tabular} &
		\begin{tabular}[c]{@{}l@{}}
			Programming\\
			language \\
			(Public code)
		\end{tabular} \\
		\midrule

		\rowcolor[HTML]{EFEFEF}
		\begin{tabular}[c]{@{}l@{}}
			DA-DRM \\
			\cite{knobloch2018distance}
		\end{tabular} &
		\begin{tabular}[c]{@{}l@{}}
			graph-based\\
			(DRM \cite{kallman2004})
		\end{tabular} &
		\begin{tabular}[c]{@{}l@{}}
			yes (no)
		\end{tabular} &
		\begin{tabular}[c]{@{}l@{}}
			2
		\end{tabular} &
		\begin{tabular}[c]{@{}l@{}}
			random shortcut\\
			post-processor
		\end{tabular} &
		\begin{tabular}[c]{@{}l@{}}
			n/a
		\end{tabular} &
		\begin{tabular}[c]{@{}l@{}}
			yes
		\end{tabular} &
		\begin{tabular}[c]{@{}l@{}}
			n/a
		\end{tabular} &
		\begin{tabular}[c]{@{}l@{}}
			ARMAR-III\\
			humanoid
		\end{tabular} &
		\begin{tabular}[c]{@{}l@{}}
			ARMAR-III\\
			humanoid
		\end{tabular} &
		\begin{tabular}[c]{@{}l@{}}
			depth camera
		\end{tabular} &
		\begin{tabular}[c]{@{}l@{}}
			3.60 GHz CPU\\
			(n/a)
		\end{tabular} &
		\begin{tabular}[c]{@{}l@{}}
			n/a\\
			(no)
		\end{tabular} \\

		\begin{tabular}[c]{@{}l@{}}
			MP-RRT\\
			\cite{zucker2007multipartite}
		\end{tabular} &
		\begin{tabular}[c]{@{}l@{}}
			SB\\
			(RRT \cite{lavalle1998rapidly})
		\end{tabular} &
		\begin{tabular}[c]{@{}l@{}}
			yes (no)
		\end{tabular} &
		\begin{tabular}[c]{@{}l@{}}
			n/a
		\end{tabular} &
		\begin{tabular}[c]{@{}l@{}}
			greedy\\
			smoothing
		\end{tabular} &
		\begin{tabular}[c]{@{}l@{}}
			spacetime
		\end{tabular} &
		\begin{tabular}[c]{@{}l@{}}
			no
		\end{tabular} &
		\begin{tabular}[c]{@{}l@{}}
			n/a
		\end{tabular} &
		\begin{tabular}[c]{@{}l@{}}
			\{2,3,4\}-DoF\\
			mobile robot
		\end{tabular} &
		\begin{tabular}[c]{@{}l@{}}
			no
		\end{tabular} &
		\begin{tabular}[c]{@{}l@{}}
			no
		\end{tabular} &
		\begin{tabular}[c]{@{}l@{}}
			n/a
		\end{tabular} &
		\begin{tabular}[c]{@{}l@{}}
			n/a\\
			(no)
		\end{tabular} \\

		\rowcolor[HTML]{EFEFEF} 
		\begin{tabular}[c]{@{}l@{}}
			RAMP\\
			\cite{vannoy2008real}
		\end{tabular} &
		\begin{tabular}[c]{@{}l@{}}
			SB \& optimi-\\
			zation-based
		\end{tabular} &
		\begin{tabular}[c]{@{}l@{}}
			yes (no)
		\end{tabular} &
		\begin{tabular}[c]{@{}l@{}}
			60
		\end{tabular} &
		\begin{tabular}[c]{@{}l@{}}
			splines \& \\
			parab. blends
		\end{tabular} &
		\begin{tabular}[c]{@{}l@{}}
			kinematic,\\
			time, energy \& \\
			manipulability
		\end{tabular} &
		\begin{tabular}[c]{@{}l@{}}
			no
		\end{tabular} &
		\begin{tabular}[c]{@{}l@{}}
			n/a
		\end{tabular} &
		\begin{tabular}[c]{@{}l@{}}
			mobile arm
		\end{tabular} &
		\begin{tabular}[c]{@{}l@{}}
			no
		\end{tabular} &
		\begin{tabular}[c]{@{}l@{}}
			no
		\end{tabular} &
		\begin{tabular}[c]{@{}l@{}}
			3.0 GHz CPU\\
			(parallel)
		\end{tabular} &
		\begin{tabular}[c]{@{}l@{}}
			C\# \& C++\\
			(no)
		\end{tabular} \\

		\begin{tabular}[c]{@{}l@{}}
			RRT$^\mathrm{X}$\\
			\cite{otte2016rrtx}
		\end{tabular} &
		\begin{tabular}[c]{@{}l@{}}
			SB\\
			(RRT* \cite{karaman2011sampling})
		\end{tabular} &
		\begin{tabular}[c]{@{}l@{}}
			yes (no)
		\end{tabular} &
		\begin{tabular}[c]{@{}l@{}}
			n/a
		\end{tabular} &
		\begin{tabular}[c]{@{}l@{}}
			n/a
		\end{tabular} &
		\begin{tabular}[c]{@{}l@{}}
			kinodynamic
		\end{tabular} &
		\begin{tabular}[c]{@{}l@{}}
			yes
		\end{tabular} &
		\begin{tabular}[c]{@{}l@{}}
			$0-40$
		\end{tabular} &
		\begin{tabular}[c]{@{}l@{}}
			\{3,4,7\}-DoF\\
			mobile robot
		\end{tabular} &
		\begin{tabular}[c]{@{}l@{}}
			no
		\end{tabular} &
		\begin{tabular}[c]{@{}l@{}}
			no
		\end{tabular} &
		\begin{tabular}[c]{@{}l@{}}
			3.40 GHz CPU\\
			(n/a)
		\end{tabular} &
		\begin{tabular}[c]{@{}l@{}}
			Julia\\
			(yes)
		\end{tabular} \\

		\rowcolor[HTML]{EFEFEF} 
		\begin{tabular}[c]{@{}l@{}}
			EBG-RRT\\ 
			\cite{yuan2020efficient}
		\end{tabular} &
		\begin{tabular}[c]{@{}l@{}}
			SB\\
			(RRT \cite{lavalle1998rapidly})
		\end{tabular} &
		\begin{tabular}[c]{@{}l@{}}
			yes (no)
		\end{tabular} &
		\begin{tabular}[c]{@{}l@{}}
			n/a
		\end{tabular} &
		\begin{tabular}[c]{@{}l@{}}
			relay node\\
			method
		\end{tabular} &
		\begin{tabular}[c]{@{}l@{}}
			time, jerk\\
			\& energy
		\end{tabular} &
		\begin{tabular}[c]{@{}l@{}}
			no
		\end{tabular} &
		\begin{tabular}[c]{@{}l@{}}
			n/a
		\end{tabular} &
		\begin{tabular}[c]{@{}l@{}}
			6-DoF\\
			Aubo-i5 arm
		\end{tabular} &
		\begin{tabular}[c]{@{}l@{}}
			no
		\end{tabular} &
		\begin{tabular}[c]{@{}l@{}}
			no
		\end{tabular} &
		\begin{tabular}[c]{@{}l@{}}
			1.80 GHz CPU\\
			(n/a)
		\end{tabular} &
		\begin{tabular}[c]{@{}l@{}}
			ROS Matlab\\
			(no)
		\end{tabular} \\

		\begin{tabular}[c]{@{}l@{}}
			STL-RT-\\
			RRT* \cite{linard2023real}
		\end{tabular} &
		\begin{tabular}[c]{@{}l@{}}
			SB \& STL\\
			(RT-RRT* \cite{naderi2015rt})
		\end{tabular} &
		\begin{tabular}[c]{@{}l@{}}
			yes (no)
		\end{tabular} &
		\begin{tabular}[c]{@{}l@{}}
			$10$
		\end{tabular} &
		\begin{tabular}[c]{@{}l@{}}
			spatio-temporal\\
			behavior
		\end{tabular} &
		\begin{tabular}[c]{@{}l@{}}
			n/a
		\end{tabular} &
		\begin{tabular}[c]{@{}l@{}}
			no
		\end{tabular} &
		\begin{tabular}[c]{@{}l@{}}
			$0.3-1.1$
		\end{tabular} &
		\begin{tabular}[c]{@{}l@{}}
			mobile robot
		\end{tabular} &
		\begin{tabular}[c]{@{}l@{}}
			Pepper\\
			mobile robot
		\end{tabular} &
		\begin{tabular}[c]{@{}l@{}}
			n/a
		\end{tabular} &
		\begin{tabular}[c]{@{}l@{}}
			1.90 GHz CPU\\
			(n/a)
		\end{tabular} &
		\begin{tabular}[c]{@{}l@{}}
			Python\\
			(yes)
		\end{tabular} \\

		\rowcolor[HTML]{EFEFEF} 
		\begin{tabular}[c]{@{}l@{}}
			RRT-ERG\\
			\cite{merckaert2024real}
		\end{tabular} &
		\begin{tabular}[c]{@{}l@{}}
			SB \& ERG\\
			(RRT \cite{lavalle1998rapidly})
		\end{tabular} &
		\begin{tabular}[c]{@{}l@{}}
			yes (no)
		\end{tabular} &
		\begin{tabular}[c]{@{}l@{}}
			$50$
		\end{tabular} &
		\begin{tabular}[c]{@{}l@{}}
			trajectory-\\
			based ERG
		\end{tabular} &
		\begin{tabular}[c]{@{}l@{}}
			kinodynamic
		\end{tabular} &
		\begin{tabular}[c]{@{}l@{}}
			yes
		\end{tabular} &
		\begin{tabular}[c]{@{}l@{}}
			n/a
		\end{tabular} &
		\begin{tabular}[c]{@{}l@{}}
			yes (n/a)
		\end{tabular} &
		\begin{tabular}[c]{@{}l@{}}
			7-DoF Panda\\
			arm			
		\end{tabular} &
		\begin{tabular}[c]{@{}l@{}}
			ZED 2 RGB-D\\
			stereo camera
		\end{tabular} &
		\begin{tabular}[c]{@{}l@{}}
			3.80 GHz CPU\\
			(parallel)
		\end{tabular} &
		\begin{tabular}[c]{@{}l@{}}
			C++\\
			(no)
		\end{tabular} \\

		\begin{tabular}[c]{@{}l@{}}
			analytical\\
			IK \cite{shao2024online}
		\end{tabular} &
		\begin{tabular}[c]{@{}l@{}}
			extended \\ analytical IK
		\end{tabular} &
		\begin{tabular}[c]{@{}l@{}}
			yes (no)
		\end{tabular} &
		\begin{tabular}[c]{@{}l@{}}
			$\sim 4000$
		\end{tabular} &
		\begin{tabular}[c]{@{}l@{}}
			splines
		\end{tabular} &
		\begin{tabular}[c]{@{}l@{}}
			kinematic
		\end{tabular} &
		\begin{tabular}[c]{@{}l@{}}
			yes
		\end{tabular} &
		\begin{tabular}[c]{@{}l@{}}
			$0.1-0.3$
		\end{tabular} &
		\begin{tabular}[c]{@{}l@{}}
			7-DoF arm
		\end{tabular} &
		\begin{tabular}[c]{@{}l@{}}
			7-DoF Kinova\\
			Gen3 arm
		\end{tabular} &
		\begin{tabular}[c]{@{}l@{}}
			Intel Real-\\
			Sense D435i
		\end{tabular} &
		\begin{tabular}[c]{@{}l@{}}
			2.50 GHz CPU\\
			(n/a)
		\end{tabular} &
		\begin{tabular}[c]{@{}l@{}}
			n/a\\
			(no)
		\end{tabular} \\

		\rowcolor[HTML]{EFEFEF} 
		\begin{tabular}[c]{@{}l@{}}
			FPGA-\\
			based \cite{murray2016robot}
		\end{tabular} &
		\begin{tabular}[c]{@{}l@{}}
			SB\\
			(PRM \cite{kavraki1996probabilistic})
		\end{tabular} &
		\begin{tabular}[c]{@{}l@{}}
			yes (no)
		\end{tabular} &
		\begin{tabular}[c]{@{}l@{}}
			$\sim 1000$
		\end{tabular} &
		\begin{tabular}[c]{@{}l@{}}
			no
		\end{tabular} &
		\begin{tabular}[c]{@{}l@{}}
			no
		\end{tabular} &
		\begin{tabular}[c]{@{}l@{}}
			no
		\end{tabular} &
		\begin{tabular}[c]{@{}l@{}}
			$0$
		\end{tabular} &
		\begin{tabular}[c]{@{}l@{}}
			Kinova Jaco-2\\
			6-DoF arm
		\end{tabular} &
		\begin{tabular}[c]{@{}l@{}}
			Kinova Jaco-2\\
			6-DoF arm
		\end{tabular} &
		\begin{tabular}[c]{@{}l@{}}
			4x Kinect-2\\
			sensors
		\end{tabular} &
		\begin{tabular}[c]{@{}l@{}}
			FPGA\\
			(parallel)
		\end{tabular} &
		\begin{tabular}[c]{@{}l@{}}
			Verilog\\
			(no)
		\end{tabular} \\

		\begin{tabular}[c]{@{}l@{}}
			CuRobo\\ 
			\cite{sundaralingam2023curobo}
		\end{tabular} &
		\begin{tabular}[c]{@{}l@{}}
			SB \& optimi-\\
			zation-based
		\end{tabular} &
		\begin{tabular}[c]{@{}l@{}}
			yes (no)
		\end{tabular} &
		\begin{tabular}[c]{@{}l@{}}
			$\sim 50$
		\end{tabular} &
		\begin{tabular}[c]{@{}l@{}}
			circular\\
			blends
		\end{tabular} &
		\begin{tabular}[c]{@{}l@{}}
			kinematic
		\end{tabular} &
		\begin{tabular}[c]{@{}l@{}}
			no
		\end{tabular} &
		\begin{tabular}[c]{@{}l@{}}
			$0$
		\end{tabular} &
		\begin{tabular}[c]{@{}l@{}}
			6-DoF UR5e \&\\ 
			UR10 arm
		\end{tabular} &
		\begin{tabular}[c]{@{}l@{}}
			6-DoF UR5e \&\\ 
			UR10 arm
		\end{tabular} &
		\begin{tabular}[c]{@{}l@{}}
			Intel Real-\\
			Sense D415
		\end{tabular} &
		\begin{tabular}[c]{@{}l@{}}
			NVIDIA RTX\\
			4090 GPU (par.)
		\end{tabular} &
		\begin{tabular}[c]{@{}l@{}}
			Python \& \\
			CUDA (yes)
		\end{tabular} \\

		\rowcolor[HTML]{EFEFEF} 
		\begin{tabular}[c]{@{}l@{}}
			MARS\\
			\cite{tonola2023anytime} 
		\end{tabular} &
		\begin{tabular}[c]{@{}l@{}}
			SB (DRRT \cite{ferguson2006replanning},\\ MP-RRT \cite{zucker2007multipartite})
		\end{tabular} &
		\begin{tabular}[c]{@{}l@{}}
			yes (no)
		\end{tabular} &
		\begin{tabular}[c]{@{}l@{}} 
			$5$
		\end{tabular} &
		\begin{tabular}[c]{@{}l@{}}
			splines 
		\end{tabular} &
		\begin{tabular}[c]{@{}l@{}}
			no 
		\end{tabular} &
		\begin{tabular}[c]{@{}l@{}}
			no 
		\end{tabular} &
		\begin{tabular}[c]{@{}l@{}}
			$0-0.5$ 
		\end{tabular} &
		\begin{tabular}[c]{@{}l@{}}
			\{6,12,18\}-DoF\\ 
			arm, 3D point
		\end{tabular} &
		\begin{tabular}[c]{@{}l@{}}
			6-DoF UR10e\\
			arm
		\end{tabular} &
		\begin{tabular}[c]{@{}l@{}}
			Intel Real-\\
			Sense D435i
		\end{tabular} &
		\begin{tabular}[c]{@{}l@{}}
			2.80 GHz CPU\\
			(parallel) 
		\end{tabular} &
		\begin{tabular}[c]{@{}l@{}}
			ROS C++\\
			(yes)
		\end{tabular} \\

		\begin{tabular}[c]{@{}l@{}}
			\textbf{DRGBT}\\
			\textbf{(ours)}
		\end{tabular} &
		\begin{tabular}[c]{@{}l@{}}
			\textbf{SB (RGBT \cite{lacevic2020gbur},}\\
			\textbf{RGBMT* \cite{covic2023asymptotically})}
		\end{tabular} &
		\begin{tabular}[c]{@{}l@{}}
			\textbf{yes (yes)} 
		\end{tabular} &
		\begin{tabular}[c]{@{}l@{}}
			$\bb{100}$
		\end{tabular} &
		\begin{tabular}[c]{@{}l@{}}
			\textbf{splines}
		\end{tabular} &
		\begin{tabular}[c]{@{}l@{}}
			\textbf{kinematic}
		\end{tabular} &
		\begin{tabular}[c]{@{}l@{}}
			\textbf{yes}
		\end{tabular} &
		\begin{tabular}[c]{@{}l@{}}
			$\bb{0-1.6}$
		\end{tabular} &
		\begin{tabular}[c]{@{}l@{}}
			\textbf{\{2,6,10,18\}-}\\ 
			\textbf{DoF arm}
		\end{tabular} &
		\begin{tabular}[c]{@{}l@{}}
			\textbf{6-DoF UFactory}\\ 
			\textbf{xArm6 arm}
		\end{tabular} &
		\begin{tabular}[c]{@{}l@{}}
			\textbf{2x Intel Real-}\\
			\textbf{Sense D435i}
		\end{tabular} &
		\begin{tabular}[c]{@{}l@{}} 
			\textbf{2.60 GHz CPU}\\
			\textbf{(sequential)}
		\end{tabular} &
		\begin{tabular}[c]{@{}l@{}}
			\textbf{ROS2 C++}\\
			\textbf{(yes)}
		\end{tabular} \\
		\bottomrule
		\vspace{-1.1cm}
	\end{tabular}%
}
\end{table*}

Dynamic roadmaps (DRM) algorithm \cite{kallman2004} analyzes trade-offs between maintaining dynamic roadmaps and applying an online bidirectional rapidly-exploring random trees (RRT-Connect) planner \cite{kuffner2000rrt}. The DRM-based real-time approach is proposed in \cite{liu2010dynamic}, which deals with unpredictable environments using both the subgoal generator and the inner replanner. Hierarchical DRM method \cite{yang2017hdrm} uses a unique hierarchical structure to efficiently utilize the information about configuration-to-workspace occupation. Distance-aware DRM algorithm \cite{knobloch2018distance} extends DRM by planning a path in DE considering the distance to obstacles. The algorithm uses a so-called \textit{voxel distance grid}, which is updated due to obtained measurements. During the roadmap search, the distance information is used within a cost function, and for trajectory smoothing in the end. 

RRT-based algorithms for DEs have been concurrently developed with those based on PRM. One of the first approaches is execution-extended RRT (ERRT) algorithm \cite{bruce2002real}, which improves replanning efficiency and the quality of generated paths by using a so-called \textit{waypoint cache memory} in $\mathcal{C}$-space. Afterward, dynamic RRT (DRRT) algorithm \cite{ferguson2006replanning}, based on the D* family of deterministic algorithms \cite{stentz1995focussed}, is proposed. It simply removes occupied edges in a tree, while collision-free ones are retained. 
Multipartite RRT (MP-RRT) algorithm \cite{zucker2007multipartite} combines the advantages of existing RRT adaptations for dynamic MP from \cite{li2002, kallman2004, ferguson2006replanning}. This is conveniently achieved by affecting a sampling distribution and reusing previously planned edges. Chance-constrained RRT \cite{luders2010chance} is an RTMP algorithm known for using so-called \textit{chance constraints} to guarantee probabilistic feasibility in DE.

\textit{Anytime} variant of DRRT is developed in \cite{ferguson2007anytime}, based on incremental sampling, that efficiently computes motion plans and improves them continually towards an optimal solution. Anytime algorithm based on RRT* \cite{karaman2011anytime} is proposed using the inspiration from RRT* \cite{karaman2011sampling}. They stand out for almost-surely asymptotically optimal methods. The first RTMP variant of RRT* is presented in \cite{naderi2015rt} as RT-RRT* algorithm equipped with an online tree rewiring strategy that allows tree roots to move with the robot without discarding previously sampled paths. 
Similarly, Multiple Parallel RRTs (MPRRT) algorithm \cite{sun2015high} concurrently runs separate RRTs on each available processor core. Moreover, online variants of RRT* (and fast marching tree (FMT*) \cite{janson2015fast} as well) are presented in \cite{chandler2017online}. 

Real-time adaptive MP (RAMP) approach \cite{vannoy2008real} has shown to be suitable for planning high DoF redundant robots in DEs. In addition, real-time optimization of trajectories may be subjected to different criteria, such as maximizing manipulability or minimizing energy and time. Extensions of RAMP are proposed in \cite{xiao2010real} and \cite{mcleod2016real}. The first one substantially extends the RAMP paradigm to using a continuum manipulator, while the second one adapts it for non-holonomic MP.

RRT$^\mathrm{X}$ \cite{otte2016rrtx} is the first asymptotically optimal SB algorithm for real-time navigation in DEs. Whenever a change in the environment is detected, the graph in free $\mathcal{C}$-space is quickly updated so that its connection with a corresponding goal subtree is achieved optimally. 

The main goal of the approach from \cite{mercy2016real} is to transform the MP problem into a smaller dimensional optimization problem, suitable to find real-time optimal motion trajectories using so-called \textit{trajectory spline parameterization}. 

Horizon-based lazy optimal RRT from \cite{chen2019horizon} demonstrates fast and efficient RTMP in DEs by using techniques such as lazy steering, lazy collision checking search tree, forward tree pruning, and sampling distribution online learning. 

Recently, probability/efficient bias-goal factor RRT (PBG-RRT and EBG-RRT) algorithms are presented in \cite{yuan2019heuristic} and \cite{yuan2020efficient}, respectively, as modifications of ERRT. They combine a heuristic and bias-goal factor, which implies fast convergence and local minima avoidance. 


The work in \cite{finean2021predicted} explores the use of so-called \textit{composite signed-distance fields} (SDF) in MP, in order to predict obstacle motions in real time. SDF is also leveraged within the recent approach from \cite{marticorena2024rmmienhancedobstacleavoidance} using neural networks to efficiently integrate a collision avoidance cost term by maximizing the total distance to obstacles during the robot's motion.

Concepts from SB methods and nonlinear control systems theory are combined into online single-query SB motion planning/replanning in pipes (PiP-X) algorithm \cite{jaffar2022pip}, which is capable to replan motions for the class of nonlinear dynamic systems working in DEs. SB methods are exploited to generate a so-called \textit{funnel-graph}, which is then used to determine an optimal path -- \textit{funnel-path} to lead the robot to the goal.

The method from \cite{liu2022collision} proposes an effective way of online generating trajectories in the robot's workspace focusing on avoiding dynamic obstacles. MP consists of two parts: \textit{front-end} path search, where an initial trajectory is generated to be both safe and feasible regarding all kinodynamic constraints; and \textit{back-end} optimization of the trajectory using cubic B-spline optimization. Finally, the obtained trajectory is transformed to $\mathcal{C}$-space using inverse kinematics.

Space-time RRT* (ST-RRT*) \cite{grothe2022st}, inspired by RRT-Connect, is a bidirectional, probabilistically complete, and asymptotically optimal MP algorithm. It can operate in unbounded time spaces with dynamic obstacles accounting for velocity constraints and unknown arrival time.

\textit{Signal temporal logic} (STL) constraints and preferences (a rigorous specification language) are incorporated into real-time RRT* \cite{linard2023real}. A cost function steering the robot towards an asymptotically optimal solution is proposed, which is one that best satisfies the STL requirements.

The RRT's inability to handle dynamic constraints, and the trajectory-based \textit{explicit reference governor's} undesirable property of getting trapped in local minima are overcome to formulate a computationally efficient planning and control architecture \cite{merckaert2024real}, which can steer the manipulator's end-effector in the presence of actuator limits, a cluttered static obstacle environment, and moving human collaborators.

The recently proposed method in \cite{shao2024online} uses geometry-based analytical inverse kinematics (IK) to guide fast collision avoidance reactive planning in real time. The algorithm is developed into a local replanner, and integrated into a global trajectory generation framework to avoid unforeseen dynamic obstacles. 

Hardware acceleration has long been used in MP. The authors in \cite{murray2016robot} developed a dedicated FPGA-based circuitry to solve the planning problem in less than $\mathrm{1\,[ms]}$ for a 6-DoF manipulator. The collision detection time for PRM edges is demonstrated to be virtually independent of the graph size. Similarly, a massive GPU acceleration has been applied in \cite{sundaralingam2023curobo} demonstrating a parallel optimization technique to solve the MP problem for manipulators. A significant speedup over state-of-the-art methods is achieved. GPU-accelerated library -- CuRobo is released, which offers a parallel geometric planner and a collision-free IK solver. However, a price to pay in both \cite{murray2016robot} and \cite{sundaralingam2023curobo} is having a dedicated, expensive hardware with a heavy GPU acceleration or specific FPGA-based circuitry.


The paper \cite{thomason2024motions} introduces a new perspective based on \textit{vector-oriented operations} that accelerates routines used by SB algorithms by over 500 times. Planning times are reduced to the range of microseconds without requiring specialized hardware (e.g., FPGA, GPUs). By leveraging fine-grained parallelism, the approach enhances critical subroutines such as forward kinematics, collision checking, and nearest neighbor search, demonstrating its effectiveness for 7--14 DoF robots.

A particular approach -- Multi-path replanning strategy (MARS) \cite{tonola2023anytime} utilizes the anytime feature to exploit a set of precomputed paths to compute a new path on a multi-thread architecture within a few hundred $\mathrm{[ms]}$ whenever a path becomes obstructed. An informed sampling is used to build an oriented graph that can reuse results from the previous replanning iterations. MARS has shown superior in terms of success rate and quality of solutions when compared to competing state-of-the-art methods. For more information about RTMP in DEs, the reader is referred to surveys in \cite{mohanan2018survey, zhou2022review, tamizi2023review}. 

We stress that all of the aforementioned planners deal with at least one issue from the list below:
\vspace{-0.05cm}
\begin{itemize}[leftmargin=0.35cm]
	\item the algorithm is validated using low-DoF scenarios (e.g., mobile robots or robotic arms with fewer than six DoFs);
	\item validation is carried out only in simulation and not using sensor-based MP in real-world scenarios; 
	\item relatively expensive hardware (perception, multiple CPUs, GPUs, etc.) is required along with the utilization of concurrent programming;
	\item the claims about real-time execution are  empirically established based on successful experiments without the explicit schedulability analysis that handles execution deadlines;
	\item algorithms typically run at relatively low frequencies;
	\item slow computation of plans, sometimes without replanning routine at all;
	\item lacking anytime feature for improving paths on the fly;
	\item relying on heavy-duty learning, data preprocessing, and/or complex data structures.
\end{itemize}

To this end, we propose a novel approach that aims at overcoming all of the aforementioned issues. Relevant features of the state-of-the-art algorithms are summarized within Tab. \ref{tab_features_comparison}. The sign ``n/a'' stands for the feature that was either not available or not possible to find despite our best efforts. 

The preliminary results of this research have been published in \cite{covic2021path}. This paper further contributes with the following:
\vspace{-0.05cm}
\begin{itemize}[leftmargin=0.35cm]
	\item scheduling framework that enables hard real-time execution of the planner in DEs;
	\item formulation of a new structure -- dynamic expanded bubble, which is exploited to guarantee a safe collision-free motion of the robot under kinematic constraints;
	\item time parameterization of collision-free local paths satisfying all imposed kinematic constraints;
	\item extensive analysis of execution time of critical routines with respect to the number of obstacles;
	\item a comprehensive comparative study revealing DRGBT outperforms competing state-of-the-art methods;
	\item sensor-based experimental validation of the proposed planning algorithm using the real robot through scenarios involving moving objects and human-robot coexistence. 
\end{itemize}

The remainder of the paper is structured as follows. Sec. \ref{Sec. DRGBT Algorithm} provides an overview of DRGBT algorithm from \cite{covic2021path}. Sec. \ref{Sec. Scheduling Framework of the Algorithm} proposes a scheduling framework for the real-time execution of the algorithm. Sec. \ref{Sec. Safe Motion of the Robot in Dynamic Environments Under Bounded Obstacle Velocity} proposes a concept of dynamic bubbles and burs, and proves that a guaranteed safe motion of the robot in DEs can be achieved. A simulation study within Sec. \ref{Sec. Simulation Study} reveals the most critical routines, and confirms that the chosen scheduling framework ensures the real-time execution of the algorithm. Moreover, a comparative study within Sec. \ref{Sec. Comparison to State-of-the-art Methods} juxtaposes DRGBT to state-of-the-art methods. Sec. \ref{Sec. Experiments} provides experimental results demonstrating the real-time execution. Finally, Sec. \ref{Sec. Discussion and Conclusions} brings some discussion, concluding remarks, and future work directions.

\begin{figure}[t]
	\centering
	\includegraphics[width=0.85\linewidth]{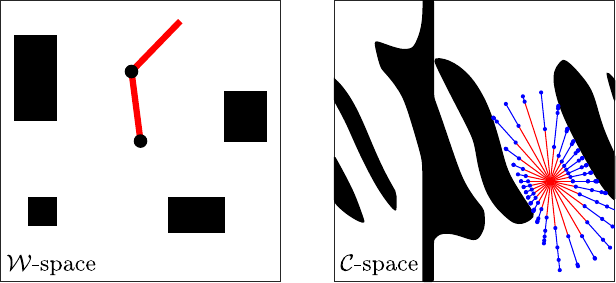}
	\vspace{-0.1cm}
	\caption{Generalized bur with 30 ``spines'' (radial directions) and 5 ``layers'' (extensions along a single spine) (right) corresponding to a configuration of 2-DoF robot with four world obstacles (left) \cite{lacevic2020gbur}.}
	\label{fig_gbur}
	\vspace{-0.6cm}
\end{figure}

\begin{figure*}[t]
\centering
\includegraphics[width=\linewidth]{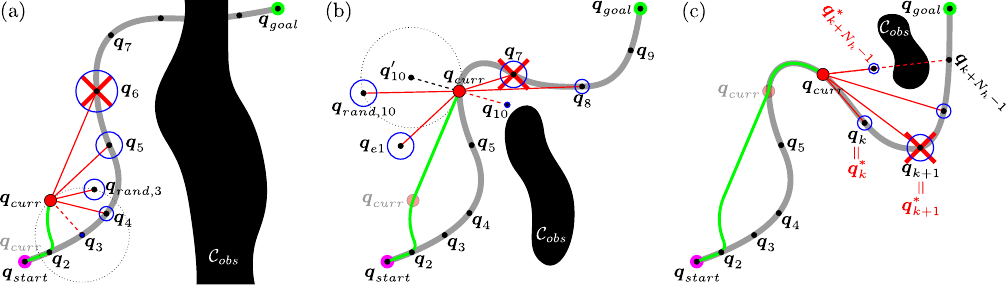}
\vspace{-0.55cm}
\caption{Graphical interpretation of some DRGBT components: obstacles (black); predefined/traversed path (gray/green); local (reached) horizon $\bb{Q}_h$ ($\color{red}\bb{Q}_h^*$) with the corresponding horizon spines, i.e., generalized bur (solid red lines); next target node (red ``$\times$''); the value of node weight (the size of the blue circle centered at the corresponding node). (a) The obstacle blocks the predefined path to the goal; (b) The obstacle has moved in a way that clears the path to the goal, thus replanning has been performed and the predefined path has been changed; (c) The process of generating new local horizon and path replanning (as needed) continues.}
\label{fig_DRGBT}
\vspace{-0.6cm}
\end{figure*}

\vspace{-0.25cm}
\section{DRGBT Algorithm}
\label{Sec. DRGBT Algorithm}
\vspace{-0.05cm}
For better comprehension, we first introduce the notation and define the problem statement in Subsec. \ref{Subsec. Problem Statement and Assumptions}. The background used within the proposed approach is described in Subsec. \ref{Subsec. Background}. Then, we broadly describe DRGBT algorithm with its main features through Subsecs. \ref{Subsec. A Brief Overview of the Algorithm} and \ref{Subsec. Anytime Feature of DRGBT}, while the details including pseudocode are given in Subsec. \ref{Subsec. The Course of the Algorithm}.

\vspace{-0.3cm}
\subsection{Problem Statement and Assumptions}
\label{Subsec. Problem Statement and Assumptions}
The \textit{robot's $n$-dimensional configuration space} is denoted as $\mathcal{C}$. The \textit{obstacle space} $\mathcal{C}_{obs} \subseteq \mathcal{C}$ is the closed subset of $\mathcal{C}$ which implies the collision with obstacles or self-collision. The \textit{free space} $\mathcal{C}_{free} = \mathcal{C} \backslash \mathcal{C}_{obs}$ is the open subset of $\mathcal{C}$ that the robot can reach. It is assumed that $\mathcal{C}_{obs}$ corresponds to a finite number $N_{obs}$ of (possibly overlapping) convex \textit{world obstacles} $\mathcal{WO}_{j}$, $j\in\{1,2,\dots,N_{obs}\}$, existing in the \textit{robot's workspace} $\mathcal{W}$. The case of non-convex obstacles can be tackled whether by using some explicit decomposition techniques (e.g., \cite{lien2008approximate}), or indirectly through algorithms for collision/distance checking that use an implicit representation of the environment via bounding volume hierarchies (BVHs), which typically rely on convex geometric primitives \cite{pan2012fcl}.

The \textit{robot's start} and \textit{goal configurations} are denoted as $\bb{q}_{start}$ and $\bb{q}_{goal}$, respectively. At any given time $t$, the \textit{robot's current configuration} is represented by $\bb{q}_{curr}(t)$, where $\bb{q}_{curr} : t\in[t_0, t_{curr}] \mapsto \mathcal{C}$ determines the \textit{robot's traversed path} from the \textit{initial time} $t_0$ to the \textit{current time} $t_{curr}$. This path is undefined for $t > t_{curr}$. Since obstacles move, occupied and free parts of $\mathcal{C}$ vary over time, i.e., $\mathcal{C}_{obs} = \mathcal{C}_{obs}(t)$ and $\mathcal{C}_{free} = \mathcal{C}_{free}(t)$. The changes in $\mathcal{W}$ must be perceived by exteroceptive sensors (e.g., depth cameras). Adequate sampling strategy should in turn capture the corresponding changes in $\mathcal{C}_{obs}(t)$ and $\mathcal{C}_{free}(t)$ to facilitate replanning when necessary.

We stress that the perception is not within the scope of this research. Moreover, our approach is completely agnostic to the system of perception as long as it provides a representation of the environment via convex geometric primitives in real time.

A \textit{motion trajectory} $\bb{\pi}(t) = \bb{\pi}[\bb{q}_0, \bb{q}_f]$ represents a curve defined by a continuous mapping $\bb{\pi} : t\in [t_0, t_f] \mapsto \mathcal{C}$ such that $\bb{\pi}(t_0) \mapsto \bb{q}_0$ and $\bb{\pi}(t_f) \mapsto \bb{q}_f$, where $t_0$ is the initial time and $t_f$ is a \textit{final time} (in general not a priori given). A trajectory is considered \textit{valid} if and only if both $\bb{\pi}(t) \cap \mathcal{C}_{obs}(t) = \varnothing$, $\forall t \in[t_0, t_f]$, and it is feasible for the robot to follow $\bb{\pi}(t)$ within the limits of its kinematic and other constraints $\mathcal{K}$.

Finally, the goal of MP in DEs is to compute a trajectory $\bb{\pi}[\bb{q}_{start}, \bb{q}_{goal}]$, where $\bb{q}_{curr}(t)\in \bb{\pi}(t)$ remains collision-free over the time $t$, ensuring that the robot navigates from $\bb{q}_{start}$ to $\bb{q}_{goal}$ while respecting dynamic changes in $\mathcal{C}_{obs}(t)$, as well as $\mathcal{K}$. More precisely, $\bb{q}_{curr}(t) \in \mathcal{C}_{free}(t)$, $\forall t\in[t_{start}, t_{goal}]$.

\vspace{-0.3cm}
\subsection{Background}
\label{Subsec. Background}
The concept of \textit{bubbles of free $\mathcal{C}$-space} was introduced in \cite{quinlan1994real} as a method for defining collision-free regions around a given configuration based on a single distance information from the workspace. A bubble $\mathcal{B}(\bb{q}, d_c)$ at configuration $\bb{q}$ is computed using the minimal distance $d_c$ between the robot and the set of obstacles $\mathcal{WO}$ within the workspace $\mathcal{W}$.


Although bubbles of free $\mathcal{C}$-space have proven to be a valuable tool for path modification and motion planning, their underlying definition is rather conservative. To satisfy convexity constraints and ensure computational simplicity, the size of these bubbles is often unnecessarily restricted. In response to this limitation, a so-called \textit{complete bubble of free $\mathcal{C}$-space} is proposed in \cite{lacevic2016burs}. Such alternative volume-based, less conservative structure still relies on the same input information as the original bubble. By intersecting $N$ rays originating from the configuration $\bb{q}$ with the complete bubble, so-called \textit{burs of free $\mathcal{C}$-space} are acquired, as depicted by red lines in Fig. \ref{fig_gbur} (right).

DRGBT algorithm, originally proposed in \cite{covic2021path}, uses the idea of \textit{generalized burs of free $\mathcal{C}$-space} from \cite{lacevic2020gbur}. This kind of bur is proven to capture relatively large portions of free $\mathcal{C}$-space, using the distance query from the workspace (Fig. \ref{fig_gbur}). This enables fast $\mathcal{C}$-space exploration and path planning, which facilitates the algorithm operation in DEs. Roughly speaking, the generalized bur resembles the set of range sensors able to perceive the free $\mathcal{C}$-space locally and thus aid the algorithm in computing quality collision-free paths. This feature turns out to be particularly helpful in replanning stages.

\vspace{-0.35cm}
\subsection{A Brief Overview of the Algorithm}
\label{Subsec. A Brief Overview of the Algorithm}
\vspace{-0.05cm}
In the sequel, all components of DRGBT will be briefly presented referring to Fig. \ref{fig_DRGBT}. The reader can find more technical details in \cite{covic2021path}. DRGBT starts with planning of initial path from $\bb{q}_{start}$ to $\bb{q}_{goal}$ using any static planner. Specifically, we use RGBT-Connect \cite{lacevic2020gbur} or RGBMT* \cite{covic2023asymptotically}. If the path is found, it is labeled \textit{predefined path}, consisting of nodes $\bb{q}_1 = \bb{q}_{start},\, \dots, \bb{q}_K = \bb{q}_{goal}$, which is then attempted to follow. Precisely, its $N_h$ nearest nodes are exploited to render a so-called \textit{local horizon} $\bb{Q}_h = \{\bb{q}_k,\,\dots,\, \bb{q}_{k+N_h-1}\}$, $k\in\{2,\dots,K-N_h+1\}$ (e.g., nodes $\bb{q}_3,\dots, \bb{q}_6$ in Fig. \ref{fig_DRGBT} (a)). If $K-N_h+1<k\leq K$, then $N_h-K+k$ random nodes, chosen from neighborhood of $\bb{q}_{curr}$, are added to the horizon in order to ensure the horizon size maintains $N_h$.
	
In case when the initial path is not found, $N_h$ random nodes are generated from the neighborhood of $\bb{q}_{start}$, which are then added to $\bb{Q}_h$. After the horizon is generated, \textit{horizon spines} (i.e., generalized bur) are computed, starting from $\bb{q}_{curr}$ towards all nodes from $\bb{Q}_h$ to obtain a so-called \textit{reached horizon} $\bb{Q}_h^* = \{\bb{q}_k^*,\,\dots,\, \bb{q}_{k+N_h-1}^*\}$ (as Fig. \ref{fig_DRGBT} (c) shows).

The next step is the quality assessment of each reached node from $\bb{Q}_h^*$, where each one is assigned a node weight (as depicted in Fig. \ref{fig_DRGBT}). This procedure is a heuristic attempt to capture two important aspects: node proximity to the goal, and the minimal workspace distance to obstacles $d_c$, along with its rate. Furthermore, all estimated node weights are used to decide whether to replan the predefined path. If yes (e.g., Fig. \ref{fig_DRGBT} (b, c)), the replanning attempts to obtain a new path from $\bb{q}_{curr}$ to $\bb{q}_{goal}$. In case the new path is not found, the replanning is automatically triggered in the next iteration, while the robot structurally explores the neighborhood of $\bb{q}_{curr}$.

It is worth noting that the horizon can be modified. For this purpose, two types of nodes are introduced: a \textit{bad node} -- a node with zero weight, and a \textit{critical node} -- a node whose $d_c$ (or its underestimate -- see \cite{lacevic2020gbur}, \cite{covic2021path}) is less than a specified threshold $d_{crit}$. Nodes of both types are replaced with possibly better random nodes, generated in their (near) neighborhood. This mechanism aims at reducing the probability of collision. For example, Fig. \ref{fig_DRGBT} (a) / (b) shows that the bad / critical node $\bb{q}_3$ / $\bb{q}_{10}$ is replaced with $\bb{q}_{rand,3}$ / $\bb{q}_{rand,10}$.

Moreover, in order to gain a more detailed insight into the neighborhood around $\bb{q}_{curr}$, the adaption of the horizon size depending on robot-obstacles proximity is provided as
\vspace{-0.15cm}
\begin{equation}\label{eq_N_h_adaptation}
	N_h = \min\left\{\left\lfloor N_{h_0} \left(1 + \frac{d_{crit}}{d_c}\right) \right\rfloor,~ n\cdot N_{h_0}\right\},
	\vspace{-0.15cm}
\end{equation}
where $N_{h_0}$ is the algorithm parameter representing an \textit{initial horizon size}. It appears beneficial to increase the horizon size when obstacles get closer to the robot \cite{covic2021path}.

In addition, so-called \textit{lateral spines} are generated in a subspace orthogonal to the current motion vector in $\mathcal{C}$-space. It was discussed in \cite{covic2021path} how the inclusion of these spines increases the likelihood of obtaining collision-free motions. 

Finally, the node with the highest weight from $\bb{Q}_h^*$ is chosen as the \textit{next target node} $\bb{q}_{next}$ at which the robot is headed (as depicted by red ``$\times$'' in Fig. \ref{fig_DRGBT}). In case there are many nodes with the same weight, the closest one to $\bb{q}_{goal}$ is chosen.

\vspace{-0.3cm}
\subsection{Anytime Feature of DRGBT}
\label{Subsec. Anytime Feature of DRGBT}
In addition to \cite{covic2021path}, here we analyze the \textit{anytime feature} of DRGBT. It strives to continuously improve the path quality minimizing its cost in $\mathcal{C}$-space while taking into account the relative value of $d_c$ along with its rate. Therefore, this subsection aims to explain such a feature regarding two aspects: horizon modification and replanning strategy.

\begin{figure}[t]
	\centering
	\includegraphics[width=0.95\linewidth]{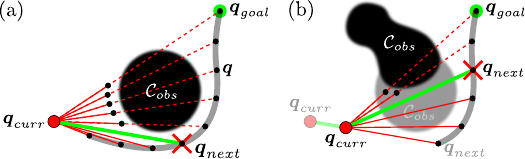}
	\caption{Generated horizon spines (solid red lines) in the current iteration (a), and in the next iteration (b). The current/previous position of the obstacle, $\bb{q}_{curr}$, and $\bb{q}_{next}$, are colored in black/gray.}
	\label{fig_DRGBT_anytime}
\end{figure}

Let us consider Fig. \ref{fig_DRGBT_anytime} that depicts DRGBT improving the current solution in an anytime manner, i.e., at each algorithm iteration. The left figure shows the generated horizon spines from $\bb{q}_{curr}$ in the current iteration. The green line indicates the current local path leading the robot towards $\bb{q}_{next}$. After the obstacle has moved in the next iteration, as shown in the right figure, the horizon is updated, as well as $\bb{q}_{next}$, thus the robot will follow the new local path (the green line). Note that three components affect a node weight computation (see \cite{covic2021path} for details): (i) a relative $d_c$, (ii) a relative change of $d_c$, and (iii) a detrended relative distance to the goal. In case the obstacle clears the configuration $\bb{q}\in\bb{Q}_h$, such that a spine $\overline{\bb{q}_{curr}\bb{q}}$ can be generated, then (i) increases and (ii) is positive, which enhance the weight of such node $\bb{q}$. After taking into account (iii), $\bb{q}$ eventually becomes a candidate for the next target node, and $\bb{q}_{next}$ may be updated. Therefore, a cost-to-go value for $\bb{q}_{curr}$ reduces, assuming that some metric function in $\mathcal{C}$-space evaluates an effort to go from $\bb{q}_{curr}$ to $\bb{q}_{goal}$ over the updated $\bb{q}_{next}$, as shown in Fig. \ref{fig_DRGBT_anytime} (b).

In case obstacles disturb or occupy certain nodes from $\bb{Q}_h$, the replanning procedure will certainly be triggered immediately or after a few iterations depending on node weights. Thus, the algorithm seeks a new, higher-quality candidate path (in terms of weights of the local horizon nodes), which reinforces the anytime flavor of DRGBT. Moreover, if some asymptotically optimal planner is employed for replanning (e.g., RRT* \cite{karaman2011sampling} or RGBMT* \cite{covic2023asymptotically}), the new path will tend to optimal one according to used criteria (e.g., path length in $\mathcal{C}$-space based on weighted Euclidean distance). Exact optimality is clearly not guaranteed since the replanning time is limited in real-time applications. However, following shortest paths in environments with changing obstacles may not be preferable in the first place, since such paths typically pass very close to obstacles. There are many ways to overcome this issue by sacrificing path-length optimality, yet increasing path safety. In DRGBT, this is taken care of by a simple heuristic function for computing node weights which depend on $d_c$.

\vspace{-0.8cm}
\subsection{The Course of the Algorithm}
\label{Subsec. The Course of the Algorithm}
\vspace{-0.25cm}
This subsection describes the flow of DRGBT approach. Some modifications are introduced with respect to the originally published algorithm \cite{covic2021path}, in order to achieve its execution in real time while keeping its high performance and proper utilization of individual routines at the same time.

The pseudocode of the algorithm is given in Alg. \ref{Pseudocode DRGBT algorithm}. Besides standard inputs required by DRGBT from \cite{covic2021path}, whose goal is to obtain collision-free geometric paths, the updated algorithm aims at computing time-parameterized trajectories and therefore has to account for constraints. To this end, we consider kinematic constraints $\mathcal{K}$ imposed on joint variables via maximal velocity $\bb{\omega}_{max}$, acceleration $\bb{\alpha}_{max}$, and (if available) jerk $\bb{j}_{max}$. The next input is real-time information about the environment, stored in $\mathcal{WO}$. The update of $\mathcal{WO}$ occurs within the function $\mathtt{perception}$ by providing the information/message about the actual pose/size of each obstacle. It is assumed that $\mathtt{perception}$ is executed by another processor in hard real-time, and, however important, is not a constitutive part of our planning algorithm. As for DRGBT's outputs, the variable $goal\_reached$ denotes whether $\bb{q}_{goal}$ is reached, and $\bb{Q}_{trav}$ is a sequence containing visited robot's configurations.

\SetEndCharOfAlgoLine{}
\begin{algorithm}[t]
	\SetAlgorithmName{Algorithm}{}
	\Indp
	\caption{DRGBT algorithm}
	\label{Pseudocode DRGBT algorithm}
	\small
	
	\KwIn{$\bb{q}_{start}$, $\bb{q}_{goal}$, $\mathcal{K}$, $\mathcal{WO}$, $safe\_on$} 
	
	\KwOut{$goal\_reached$, $\bb{Q}_{trav}$} 
	
	$\color{red}\bb{Q}_{pred} \gets \mathtt{replan}(\bb{q}_{start},\, \bb{q}_{goal})$ \\
	
	$\bb{Q}_{trav} \gets \bb{q}_{start}$ \\
	
	$\bb{q}_{curr} \gets \bb{q}_{start}$ \\
	
	$replanning \gets \False$ \\
	
	$status \gets \Reached$ \\
	
	\While{$\bb{q}_{curr}\, \neq \,\bb{q}_{goal}$}
	{
		$\color{gray}\mathcal{WO} \gets \mathtt{perception}(\bb{q}_{curr})$ \label{Alg1 perception}\\
		
		\If{$status \neq \Advanced$}
		{
			$\color{blue}\bb{Q}_h \gets  \mathtt{generateHorizon}(\bb{q}_{curr},\, \bb{Q}_{pred},\, status)$ \label{Alg1 generateHorizon} \\
		}
		
		$\color{red}\bb{d} \gets \mathtt{computeDistances}(\bb{q}_{curr},\, \mathcal{WO})$ \\
		
		$\color{blue}\bb{Q}_h \gets \mathtt{updateHorizon}(\bb{Q}_h,\, \bb{d},\, \bb{Q}_{pred})$ \\
		
		$\color{red}GBur \gets \mathtt{generateGBur}(\bb{q}_{curr},\, \bb{Q}_h,\, \bb{d})$ \\
		
		$\bb{w} \gets \mathtt{computeNodeWeights}(GBur)$ \label{Alg1 computeNodeWeights} \\
		
		$\bb{q}_{next},\, status \gets \mathtt{getNextState}(GBur,\, \bb{w})$ \\
		
		$\color{blue}\bb{q}_{curr},\, \bb{\pi}_{curr},\, status \gets \mathtt{updateCurrState}(\bb{q}_{curr},\, \bb{q}_{next},\, \mathcal{K},\, safe\_on)$ \label{Alg1 updateCurrState} \\
		
		$\bb{Q}_{trav} \gets [\bb{Q}_{trav},\, \bb{q}_{curr}]$ \label{Alg1 Q_trav} \\
		
		\If{\Not{$\mathtt{isValid}(\bb{\pi}_{curr},\, \mathcal{WO})$}} 
		{
			$goal\_reached \gets \False$ \\
			
			\Return \label{Alg1 return} \\
		}
		
		\If{$replanning$ \Or $\mathtt{whetherToReplan}(\bb{w})$} 
		{\label{Alg1 whetherToReplan}
			
			$\color{red}\bb{Q}_{pred,new} \gets \mathtt{replan}(\bb{q}_{curr},\, \bb{q}_{goal})$ \label{Alg1 replan} \\
			
			\If{$\bb{Q}_{pred,new}\,\neq\, \varnothing$} 
			{
				$\bb{Q}_{pred} \gets \mathtt{updatePath}(\bb{Q}_{pred,new},\, \bb{q}_{curr})$ \\
				
				$replanning \gets \False$ \\
				
				$status \gets \Reached$ \\
			}
			\Else
			{
				$replanning \gets \True$ \label{Alg1 replanning true} \\
			}
		}
	}
	
	$goal\_reached \gets \True$ \\
\end{algorithm}
\setlength{\textfloatsep}{0pt}

First, the predefined path $\bb{Q}_{pred}$ can be obtained by any static planner (e.g., RRT-Connect \cite{kuffner2000rrt}, RGBT-Connect \cite{lacevic2020gbur}, or RGBMT* \cite{covic2023asymptotically}). At the beginning, $\bb{Q}_{trav}$ only contains $\bb{q}_{start}$. A local variable $replanning$ denotes whether the replanning is required, and it is initially set to \False. Moreover, there is a local variable $status$ indicating the robot's current status (\Reached, \Advanced, \Trapped~-- inherited from RRT \cite{lavalle1998rapidly}).

Lines \ref{Alg1 perception}--\ref{Alg1 replanning true} are being executed, until $\bb{q}_{curr}$ reaches $\bb{q}_{goal}$ or the collision occurs. At the beginning, the function $\mathtt{generateHorizon}$ computes a horizon $\bb{Q}_h$ using $\bb{q}_{curr}$, $\bb{Q}_{pred}$, and $status$. For the computation of generalized bur, we use the generalization of $d_c$, as proposed in \cite{ademovic2016path}. For convenience, we utilize constant values $d_i$ for $i$-th separate robot's link, $i\in\{1,2,\dots,n\}$, which are acquired by the $\mathtt{computeDistances}$ function within a vector $\bb{d} = [d_1,d_2,\dots,$ $d_n]^T$. Such piecewise constant function can be easily obtained as a byproduct of computing the minimal distance $d_c = \min\{\bb{d}\}$, since distances for each pair link-obstacle $d_{i,j}$, $j\in\{1,2,\dots,N_{obs}\}$, are computed separately anyway. Therefore, $d_i = \min\{d_{i,1}, d_{i,2}, \dots, d_{i,N_{obs}}\}$.

Afterward, $\bb{Q}_h$ is updated within the $\mathtt{updateHorizon}$ routine, including the update of $N_h$ according to \eqref{eq_N_h_adaptation}. For the generation of random and lateral spines, $\bb{q}_{next}$ is required. In case it is not determined, the first node from $\bb{Q}_h$ is chosen as $\bb{q}_{next}$. A generalized bur $GBur$, starting in $\bb{q}_{curr}$ and oriented towards all nodes from $\bb{Q}_h$, is generated by the function $\mathtt{generateGBur}$. Thereafter, the $\mathtt{computeNodeWeights}$ routine computes all bur node weights $\bb{w}$. The $\mathtt{getNextState}$ function returns $\bb{q}_{next}$ based on previously obtained weights. In case all $\bb{Q}_h$ nodes are critical ones, $status$ becomes \Trapped, suggesting that $\bb{q}_{next}$ equals $\bb{q}_{curr}$. Consequently, the robot will eventually stop (which will be elaborated in detail in Sec. \ref{Sec. Safe Motion of the Robot in Dynamic Environments Under Bounded Obstacle Velocity}), and $replanning$ is set to \True to trigger the search for better nodes.

Since the robot should advance towards $\bb{q}_{next}$ by reaching a new configuration $\bb{q}_{new}$, a time parameterization of a local path $\overline{\bb{q}_{curr}\bb{q}_{next}}$ is required regarding $\mathcal{K}$. As a result, a trajectory $\bb{\pi}_{curr}(t) = \bb{\pi}[\bb{q}_{curr},\bb{q}_{next}]$ containing $\bb{q}_{new}$ is obtained, thus $\bb{q}_{curr}$ is updated to $\bb{q}_{new}$ extending $\bb{Q}_{trav}$ by line \ref{Alg1 Q_trav}. This procedure is computed within the $\mathtt{updateCurrState}$ function depending on the input parameter $safe\_on$, which indicates whether the safe variant of DRGBT, so-called DRGBT-safe (see Sec. \ref{Sec. Safe Motion of the Robot in Dynamic Environments Under Bounded Obstacle Velocity}) is used for computing $\bb{\pi}_{curr}(t)$. The status of this motion is tracked via $status$. When it becomes different from \Advanced, line \ref{Alg1 generateHorizon} is executed again. If $status$ is \textbf{reached}, it means that $\bb{q}_{next}$ is reached, thus $\bb{Q}_h$ will be updated with the new nodes from $\bb{Q}_{pred}$. Otherwise, when $status$ is \Trapped, $\bb{Q}_h$ needs to be generated using only random nodes. 

Depending on the variable $replanning$ and the function $\mathtt{whetherToReplan}$, it is decided whether the replanning occurs. Using any static planner, the path is replanned from $\bb{q}_{curr}$ to $\bb{q}_{goal}$. When a new path $\bb{Q}_{pred,new}$ is found, the function $\mathtt{updatePath}$ replaces the previous one, and $replanning$ is set to \False. Moreover, $status$ becomes \textbf{reached} in order to generate a new horizon $\bb{Q}_h$ by line \ref{Alg1 generateHorizon}. Conversely, if the new path cannot be found, $replanning$ is set to \True, thus its replanning is explicitly required during the next iteration. It is worth mentioning that $\mathtt{updatePath}$ also refines $\bb{Q}_{pred,new}$ by modifying its all nodes such that the (weighted) Euclidean distance between each two adjacent nodes is not greater than $\|\bb{\omega}_{max}\|\cdot T$, which represents maximal distance the robot can cover in $\mathcal{C}$-space during an iteration time $T$. The refined path remains the same geometrically, yet only reallocates its nodes to the carefully chosen locations. It turned out that such pre-processing yields better performance of the overall algorithm.

Finally, the function $\mathtt{isValid}$ accounts for the motion of both the robot and obstacles simultaneously, while discretely checking the motion validity. If the robot collides with obstacles, the algorithm terminates reporting failure. Otherwise, the next iteration can be executed.

\vspace{-0.3cm}
\section{Scheduling Framework of the Algorithm}
\label{Sec. Scheduling Framework of the Algorithm}
The results from our prior work \cite{covic2021path} clearly indicate that the original form of DRGBT is unsuitable for real-time execution, as its iteration time $T$ fluctuates significantly and heavily depends on the scenario. Moreover, $T$ might be infinite theoretically (e.g., if the replanner cannot find a solution at all). This variability motivated us to introduce a proper scheduling, supported by schedulability analysis and corresponding conditions for hard real-time operation. If we assume that the environment is perceived with the frequency $f_{perc} = 1/T_{perc}$, then it is possible to miss new perception measurements in case $T > T_{perc}$. Yet, such measurements could be crucial for the robot to avoid collision. Therefore, this calls for introducing proper scheduling which limits $T$, i.e., some scheduling technique must be applied to the algorithm.

A key requirement of real-time systems is that the used scheduler produces a fully predictable task execution sequence at each time instance. Accordingly, this section is mainly concerned with how DRGBT's tasks can be properly handled by a scheduler. Commonly used approaches to real-time scheduling follows clock-driven, priority-driven, and round-robin schemes \cite{liu2000real}. Recently, priority-driven methods, such as earliest deadline first (EDF), rate-monotonic scheduling (RMS), fixed priority scheduling (FPS), dynamic priority scheduling (DPS), and their variants have been particularly popular. They are applied in many real-time operating systems due to their simplicity and predictability. Considering a wide variety of available scheduling algorithms and the corresponding theoretical analysis of schedulability (\cite{liu2000real, laplante2004real, mall2009real} and \cite{mcnaughton2011parallel}), EDF method turned out to be the most convenient and will be applied thereafter. 

Consider $N$ periodic tasks with periods $T_i$, relative deadlines $D_i$, and a worst-case/maximum execution times $e_i$, for $i\in\{1,2,\dots,N\}$. Under EDF scheduling, assuming all tasks are independent and preemptable, sufficient condition which guarantees that all tasks will meet their deadlines on a single-core processor is given as (see \cite{liu2000real} for proof)
\vspace{-0.2cm}
\begin{equation}\label{eq_utilization_of_tasks_general}
	\sum_{i=1}^{N} \frac{e_i}{\min \{D_i, T_i\}} \leq 1.
	\vspace{-0.2cm}
\end{equation}
This inequality guides the selection of task timing constraints. To accommodate the algorithm for schedulability analysis, it is decomposed into two main tasks described in the sequel.

\begin{figure}[t]
	\centering
	\includegraphics[width=\linewidth]{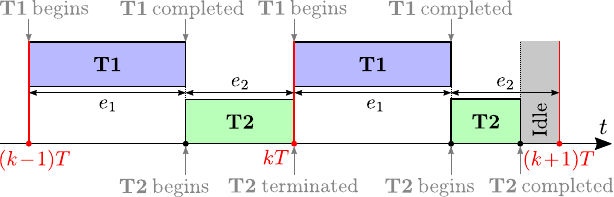}
	\vspace{-0.5cm}
	\caption{EDF scheduling of \textbf{T1} and \textbf{T2} with two priority levels.}
	\label{fig_FPS_scheduling}
\end{figure}

\vspace{-0.3cm}
\subsection{Task 1: Computing the Robot's Next Configuration}
\label{Subsec. Task 1: Computing the Robot's Next Configuration}
Task 1 (\textbf{T1}) is defined as a strictly periodic \textit{hard real‐time task} with a period $T$ (identical to the algorithm iteration time) and worst‐case/maximum execution time $e_1$. \textbf{T1} is released/activated at $t = kT$, $k \in \mathbb{N}_{0}$, as depicted in Fig. \ref{fig_FPS_scheduling}. Upon each release, \textbf{T1} deterministically executes the sequence of subtasks corresponding to lines \ref{Alg1 perception}--\ref{Alg1 return} from Alg. \ref{Pseudocode DRGBT algorithm}. To ensure continuous collision avoidance, \textbf{T1} must issue control commands at a maximum interval of $T$, thereby imposing a relative deadline $D_1 = T$. 

The new target configuration $\bb{q}_{new}\in\bb{\pi}[\bb{q}_{curr},\bb{q}_{next}]$ is carefully computed via $\mathtt{updateCurrState}$ routine to guarantee reachability of $\bb{q}_{new}$ within $T$. The collision‐free property of the resulting trajectory $\bb{\pi}[\bb{q}_{curr},\bb{q}_{new}]$ over each iteration is examined in Sec. \ref{Sec. Safe Motion of the Robot in Dynamic Environments Under Bounded Obstacle Velocity}.



\vspace{-0.3cm}
\subsection{Task 2: Replanning the Predefined Path}
\label{Subsec. Task 2: Replanning the Predefined Path}
Task 2 (\textbf{T2}) is characterized as a sporadic task since it is triggered by both the variable $replanning$ and the function $\mathtt{whetherToReplan}$ (line \ref{Alg1 whetherToReplan} in Alg. \ref{Pseudocode DRGBT algorithm}). It is wrapped in a sporadic server, which operates periodically in synchrony with \textbf{T1} (with the same period $T$). \textbf{T2} has the same activation schedule as \textbf{T1}, i.e., in $t=kT$, and its relative deadline is $D_2$, which is equal to $D_1 = T$. Due to the same release times and same relative deadlines, jobs (instances) of both tasks share the same priority since the sequences of the absolute deadlines for their jobs (instances) are identical. However, EDF scheduler lets \textbf{T1} be executed always first giving it a top-level priority. Post-\textbf{T1}, the server checks line \ref{Alg1 whetherToReplan} from Alg. \ref{Pseudocode DRGBT algorithm}, and, if needed, \textbf{T2} must be executed by lines \ref{Alg1 replan}--\ref{Alg1 replanning true} during the time
\vspace{-0.2cm}
\begin{equation}\label{eq_time_execution_task2}
	e_2 = T - e_1,
	\vspace{-0.2cm}
\end{equation}
which stems from its dependency on $e_1$. More precisely, \textbf{T2} executes either until completion or until its allocated time slice $e_2$ expires, as illustrated by the clock interrupts (red lines) in Fig. \ref{fig_FPS_scheduling}. If \textbf{T2} does not complete by its deadline, it is forcibly terminated at the subsequent release of \textbf{T1} (i.e., at $t = kT$). Such termination guarantees that \textbf{T2} never exceeds its assigned execution budget $e_2$. If \textbf{T2} is terminated, the new path is not found, and replanning must be explicitly re-invoked in subsequent iterations. Nevertheless, from a scheduling perspective, \textbf{T2} is correctly completed. The rationale behind this priority assignment is further detailed in Subsec. \ref{Subsec. Imposing Time Constraints}.

\textbf{T2} is classified as a \textit{hard real-time task} for two critical reasons. First, a replanning request indicates a potential emergency situation requiring an immediate new path to maintain safety. Second, if \textbf{T2} misses deadline, it delays the release of \textbf{T1}. Accordingly, in cases where \textbf{T2} fails to generate a new path within $e_2$ (though a feasible solution might exist), the robot may rely on the previous path. Additionally, the adaptive horizon, which emulates range sensing in $\mathcal{C}$-space, provides local collision avoidance and can ultimately initiate an emergency stopping (see further details in Sec. \ref{Sec. Safe Motion of the Robot in Dynamic Environments Under Bounded Obstacle Velocity}).

To ensure both tasks meet their deadlines in hard real-time, the sufficient condition \eqref{eq_utilization_of_tasks_general}, for $D_1 = D_2 = T$, simplifies to
\vspace{-0.15cm}
\begin{equation}\label{eq_utilization_of_tasks}
	\frac{e_1}{\min \{D_1, T\}} + \frac{e_2}{\min \{D_2, T\}} \leq 1 \implies
	\frac{e_1}{T} + \frac{e_2}{T} \leq 1,
	\vspace{-0.15cm}
\end{equation}
where $u_i = \frac{e_i}{T}$ denotes the \textit{utilization of Task $i$}. Substituting \eqref{eq_time_execution_task2} into \eqref{eq_utilization_of_tasks} yields $1\leq 1$, proving schedulability for this setup.

It is important to emphasize that the schedulability analysis is primarily focused on \textbf{T1}, as execution times of its subtasks are accurately characterized and predictable under worst-case conditions (as it will be revealed by Figs. \ref{fig_routines_CDF} and \ref{fig_routines} in Sec. \ref{Sec. Simulation Study}). Once the deadline for \textbf{T1} is rigorously satisfied, the temporal constraints for \textbf{T2} arise as a consequence by \eqref{eq_time_execution_task2}.

\vspace{-0.3cm}
\section{Safe Motion of the Robot in Dynamic Environments Under Bounded Obstacle Velocity}
\label{Sec. Safe Motion of the Robot in Dynamic Environments Under Bounded Obstacle Velocity}
In case of human presence in the robot's vicinity, it is generally undesirable to unconditionally aim at fast, minimum-time trajectories. Thus, we upgrade our approach (for generating trajectories within the $\mathtt{updateCurrState}$ routine from Alg. \ref{Pseudocode DRGBT algorithm}) in terms of automatically scaling the robot's velocity in order to guarantee a safe motion of the robot. The key factor for this upgrade is the information about the \textit{minimal robot-obstacle distance} $d_c$. In particular, the robot should be able to controllably decelerate to zero speed without causing a collision with the environment\footnote{This feature is consistent with the \textit{stop category 2} defined in the standard \textit{ISO 10218 Robots and robotic devices -- Safety requirements for industrial robots}. Our approach is also motivated by the \textit{speed and separation monitoring} (SSM) concept (described in the standard \textit{ISO/TS 15066 Robots and robotic devices -- Collaborative robots}).}. The collision is allowed to occur once the robot has stopped, which will be referred to as the \textit{type II collision}, differing from the \textit{type I collision} that assumes the robot-obstacle contact with the robot having non-zero speed.

To formulate safety requirements consistent with the mechanisms of DRGBT algorithm, we first define a structure so-called \textit{dynamic expanded bubble ($\mathcal{DEB}$) of free $\mathcal{C}$-space}, as a generalization of bubbles from \cite{quinlan1994real}, \cite{ademovic2016path} and \cite{lacevic2020gbur}.

%

\vspace{-0.3cm}
\subsection{Dynamic Expanded Bubbles and Burs}
\label{Subsec. Dynamic Expanded Bubbles and Burs}

\begin{definition}\cite{ademovic2016path}
	The curve $\bb{p}(s) : [0,L] \mapsto \mathbb{R}^3$ determines a wired model of the manipulator with its total length $L$, and the curve’s natural parameter $s$. The distance-to-obstacles profile function $d(s)$ computes a minimal distance between $\bb{p}(s)$ and $\mathcal{WO}$, i.e., it assigns each point $\bb{p}(s)$ on the kinematic chain a minimal distance to $\mathcal{WO}$. As such, $d(s)$ represents a generalization of $d_c$ between the whole robot and $\mathcal{WO}$.
\end{definition}
\vspace{-0.1cm}

As mentioned in Subsec. \ref{Subsec. The Course of the Algorithm}, we use the piecewise constant approximation of the profile $d(s)$ via constant values $d_i$ for $i$-th separate robot's link, which are represented by the vector $\bb{d} = [d_1,d_2,\dots, d_n]^T$, as proposed in \cite{ademovic2016path}. Furthermore, the points $\bb{R}_{i,j}$ and $\bb{O}_{i,j}$ are obtained as a byproduct of computing $d_c$, which determine the nearest points between $i$-th robot's link and $\mathcal{WO}_{j}$, respectively. These points uniquely define the plane $\mathcal{P}_{i,j}$ (e.g., $\mathcal{P}_{i,j}(t)$ for $t=t_0$ from Fig. \ref{fig_dynamic_gbur} (b)) that conveniently divides the workspace $\mathcal{W}$ into two halfspaces, free one containing $i$-th robot's link, and the one occupied by $\mathcal{WO}_{j}$, as proven in \cite{lacevic2020gbur}. All the planes (corresponding to link-obstacle pairs) can be expressed analytically and their parameters conveniently structured within a matrix $\bb{\mathcal{P}}$. 

It is worth stressing that the algorithm assumes that obstacles can move unpredictably in any direction during each iteration as long as the magnitude of their velocities does not exceed a \textit{maximal obstacle velocity} $v_{obs}$. We do not assume any bounds on obstacles' acceleration, thus, in theory, it can be infinite. Theoretically, it is allowed for the new obstacles to suddenly emerge during the current iteration in a region $\mathcal{W}_{occ}\subset\mathcal{W}$ obtained as a union of the occupied halfspaces w.r.t. each $\mathcal{P}_{i,j}(t)$. The sudden appearance of new obstacles in the \textit{safe region} $\mathcal{W}_{safe} = \mathcal{W}\backslash \mathcal{W}_{occ}$ is not allowed.

\vspace{-0.1cm}
\begin{definition}
	Extended configuration $\widetilde{\bb{q}} \in \mathbb{R}^{(m+1)n \times 1}$ represents a concatenated vector $\left[\bb{q}^T, \dot{\bb{q}}^T, \dots, (\bb{q}^{(m)})^T \right]^T$ containing robot's joint position $\bb{q}$, joint velocity $\dot{\bb{q}}$, and other higher-order derivatives of $\bb{q}$, all up to the order $m$.
\end{definition}
\vspace{-0.1cm}

The choice of $m$ depends on the method used for computing the trajectory, mostly in terms of the order of used polynomials. For instance, we use $m=2$, and the rationale behind this will be revealed later in Subsec. \ref{Subsec. Generating Trajectories}.

\vspace{-0.1cm}
\begin{definition}
	For a given extended configuration $\widetilde{\bb{q}}$, a vector of corresponding distances $\bb{d}$, and a maximal obstacle velocity $v_{obs}$, a dynamic expanded bubble of free $\mathcal{C}$-space for a manipulator with $n$ revolute joints is defined as
	\vspace{-0.2cm}
	\begin{equation}\label{eq_dynamic_expanded_bubble}
		\mathcal{DEB}(\widetilde{\bb{q}}, \bb{d}, v_{obs}) = \left\{\bb{y}(\widetilde{\bb{q}}, t) ~\Big|~ \overline{\rho}_i(\widetilde{\bb{q}}, t) + v_{obs} t \leq d_i \right\},
		\vspace{-0.2cm}
	\end{equation}
	for $i \in\{1,2,\dots, n\}$ and $t\in\left[0, t^*\right]$, where $\overline{\rho}_i(\widetilde{\bb{q}}, t^*) + v_{obs} t^* = d_i$, $\forall i$. The sum $\overline{\rho}_i(\widetilde{\bb{q}},t) = \sum_{j=1}^{i} r_{i,j} \left|y_j(\widetilde{\bb{q}}, t)-q_j\right|$ determines a conservative upper bound on the displacement of any point on the manipulator $\mathcal{R}_i$ when changing its configuration from $\bb{q}$ to an arbitrary configuration $\bb{y}(\widetilde{\bb{q}}, t)$ during the time $t$. The weight $r_{i,j}$ represents a radius of the cylinder coaxial with $j$-th joint enclosing all parts of the robot $\mathcal{R}_i$ starting from $j$-th joint when the robot is at the configuration $\bb{q}$. The term $\mathcal{R}_i$ designates the ``incomplete'' robot starting at its base and ending up at its $i$-th joint.
\end{definition}
\vspace{-0.1cm}

\begin{figure}[t]
	\centering
	\includegraphics[width=0.9\linewidth]{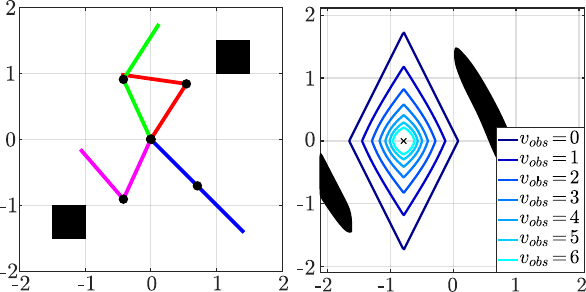}
	\caption{Workspace of a planar 2-DoF robot in four different configurations with two box-shaped obstacles (left);
		Dependence of $\mathcal{DEB}$s in $\{q_1,q_2\}$-plane on different values of $v_{obs}$ (right).}
	\label{fig_dynamic_exp_bubbles_example}
	\vspace{-0.5cm}
\end{figure}

\begin{figure}[t]
	\centering
	\includegraphics[width=0.82\linewidth]{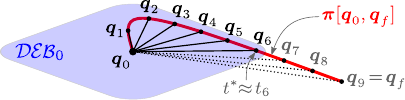}
	\caption{Process of computing dynamic bur (solid black lines).}
	\label{fig_dynamic_bur}
	\vspace{-0.1cm}
\end{figure}

Generally, any motion within $\mathcal{DEB}$ means that no point on the robot $\mathcal{R}_i$ will move more than $d_i - v_{obs} t$ for a given time $t$, thus no collision can occur. Such a defined bubble differs from the one from \cite{ademovic2016path} in terms of reducing the initial value of each $d_i$ for $v_{obs} t$. The rationale is that each $\mathcal{WO}_{j}$, may cover the distance of $v_{obs} t$ in the worst-case scenario, i.e., it may approach the robot's $i$-th link at its highest speed $v_{obs}$. This observation is equivalent to considering the plane $\mathcal{P}_{i,j}$ which artificially enlarges $\mathcal{WO}_{j}$ by following a velocity vector $\bb{v}_{\mathcal{P}_{i,j}} = \frac{\bb{R}_{i,j}-\bb{O}_{i,j}}{\|\bb{R}_{i,j}-\bb{O}_{i,j}\|} v_{obs}$ (as depicted in Fig. \ref{fig_dynamic_gbur} (b, c, d)). It is worth mentioning that we assume unpredictable motion of obstacles without monitoring or predicting their directions of motion. Otherwise, it would reduce the magnitude of the approaching vector $\bb{v}_{\mathcal{P}_{i,j}}$ from $v_{obs}$ to $v_{obs}\cdot \cos\varangle (\bb{v}_{\mathcal{P}_{i,j}}, \bb{v}_j)$, where $\bb{v}_j$ represents a monitored $j$-th obstacle velocity vector.

Fig. \ref{fig_dynamic_exp_bubbles_example} illustrates the concept of dynamic bubbles. The workspace of a planar 2-DoF robot in four different configurations with two box-shaped obstacles is shown in Fig. \ref{fig_dynamic_exp_bubbles_example} (left). Changing $v_{obs}$ from $0$ to $6\,\mathrm{[\frac{m}{s}]}$ affects the $\mathcal{DEB}$ with a root in $\bb{q}_0 = \left[-\frac{\pi}{4}, 0\right]^T$ (blue configuration) is revealed by Fig. \ref{fig_dynamic_exp_bubbles_example} (right). Note that the case for $v_{obs}=0$ corresponds with the expanded bubble from \cite{ademovic2016path}. For higher velocities, $\mathcal{DEB}$s get smaller, and their shape deviates from the diamond-like one. The initial velocity and acceleration used for time parameterization of trajectories are set to zero in all cases. 

First, in order to use \eqref{eq_dynamic_expanded_bubble}, a local path $\overline{\bb{q} \bb{y}}$, which is a straight-line segment (i.e., \textit{spine} \cite{lacevic2016burs}), must be time-parameterized since $\bb{y} = \bb{y}(\widetilde{\bb{q}}, t)$, obtained for the specific value of $t$, is a time-dependent configuration, where $\bb{y}(\widetilde{\bb{q}}, t)$ represents a family of all possible local trajectories $\bb{\pi}(t) = \bb{\pi}[\bb{q}, \bb{y}(\widetilde{\bb{q}}, t^*)]$. Computing the intersection of the specific trajectory $\bb{\pi}(t)$ with the border of $\mathcal{DEB}$ is obtained by solving the system of equations $\sum_{j=1}^{i} r_{i,j} \left|\pi_j(t)-q_j\right| + v_{obs} t = d_i$, $\forall i\in\{1,2,\dots,n\}$. Multiple solutions may exist due to the nonlinear nature of $\bb{\pi}(t)$ (e.g., polynomial). Clearly, the smallest value of $t=t^*$, where $t^*\in (0, \frac{d_c}{v_{obs}})$, should be computed. Since it could be relatively computationally expensive, we do not seek the exact value of $t^*$. Instead, we resort to a convenient workaround described in the sequel within Alg. \ref{Pseudocode Dynamic bur}. The procedure generates a so-called \textit{dynamic bur} $DBur$ using the following steps:

\begin{figure*}[t]
	\centering
	\includegraphics[width=\linewidth]{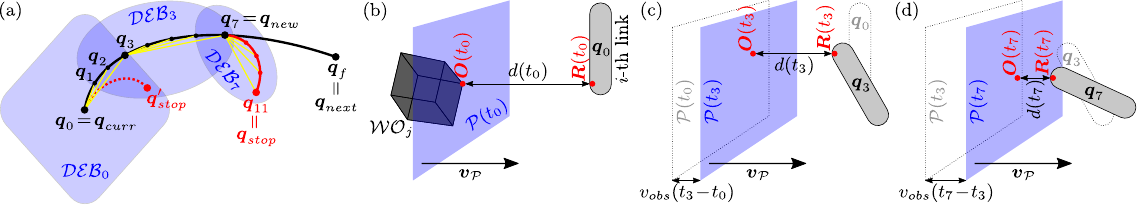}
	\vspace{-0.5cm}
	\caption{(a) Process of computing dynamic generalized bur (solid yellow lines); 
		(b,c,d) The nearest points, $\color{red}\bb{R}(t) \equiv \bb{R}_{i,j}(t)$ and $\color{red}\bb{O}(t) \equiv \bb{O}_{i,j}(t)$, from $i$-th robot's link and $\mathcal{WO}_{j}$, with the corresponding plane $\color{blue}\mathcal{P}(t) \equiv\mathcal{P}_{i,j}(t)$ and the distance $d(t)\equiv d_{i,j}(t)$ after $t\in\{t_0,t_3,t_7\}$.}
	\label{fig_dynamic_gbur}
	\vspace{-0.7cm}
\end{figure*}

\SetEndCharOfAlgoLine{}
\begin{algorithm}[t]
	\SetAlgorithmName{Algorithm}{}
	\Indp
	\caption{Dynamic bur -- $\mathtt{computeDBur}$}
	\label{Pseudocode Dynamic bur}
	\small
	
	\KwIn{$\bb{\pi}(t) = \bb{\pi}[\bb{q}_0,\bb{q}_f]$, $\Delta t$, $\bb{d}(t_0)$, $v_{obs}$}
	
	\KwOut{$DBur$} 
	
	$\bb{Q},\, DBur \gets \varnothing$ \label{Alg2 begin} \\
	
	$k \gets 0$ \\
	
	\While{$k\Delta t < \mathtt{getTime}(\bb{q}_f)$}
	{
		$\bb{Q} \gets [\bb{Q},\, \bb{\pi}(k\Delta t)]$ \\
		
		$k \gets k+1$ \\
	}
	
	$\bb{Q} \gets [\bb{Q},\, \bb{q}_f]$ \label{Alg2 line6} \\
	
	$\mathcal{DEB}_0 \gets \mathcal{DEB}(\widetilde{\bb{q}}_0,\, \bb{d}(t_0),\, v_{obs})$ \label{Alg2 DEB} \\
	
	\For{$k = 1 : \mathtt{size}(\bb{Q})$}
	{\label{Alg2 for}
		
		$t_k \gets \mathtt{getTime}(\bb{q}_k)$ \\
		
		\If{$v_{obs} (t_k-t_0) \prec \bb{d}(t_0)$ \And $\bb{q}_k \in \mathcal{DEB}_0$}
		{
			$DBur \gets [DBur,\, \overline{\bb{q}_0\bb{q}_k}]$ \\
		}
		\Else
		{
			\Return \label{Alg2 end} \\
		}
		
	}
\end{algorithm}
\setlength{\textfloatsep}{0pt}

\textbf{Step 1} (lines \ref{Alg2 begin}--\ref{Alg2 line6}): Suppose that a local trajectory $\bb{\pi}(t) = \bb{\pi}[\bb{q}_0,\bb{q}_f]$ for $t\in[t_0,t_f]$ is generated (as shown in Fig. \ref{fig_dynamic_bur}). First, a time discretization of $\bb{\pi}(t)$ is performed using a time step $\Delta t$ in order to obtain $N$ intermediate configurations $\bb{q}_{k} = \bb{\pi}(k\Delta t)$, $k\in\{1,2,\dots, N\}$, where $\bb{q}_N = \bb{q}_f$, (e.g., nodes $\bb{q}_1, \dots, \bb{q}_{9}$ in Fig. \ref{fig_dynamic_bur}). These nodes are stored in the set $\bb{Q}$. 

\textbf{Step 2} (line \ref{Alg2 DEB}): The bubble $\mathcal{DEB}(\widetilde{\bb{q}}_0, \bb{d}(t_0), v_{obs}) \equiv \mathcal{DEB}_0$ with a root in $\bb{q}_0$ is computed using the initial values for $\bb{d} = \bb{d}(t_0)$ obtained at $\bb{q}_0$. The computation of $\mathcal{DEB}_0$ requires new enclosing radii $r_{i,j}$, $j\in\{1,2,\dots,i\}$, which are easily obtained by a single forward kinematics computation at $\bb{q}_0$. 

\textbf{Step 3} (lines \ref{Alg2 for}--\ref{Alg2 end}): The idea for computing $DBur_0$ with a root in $\bb{q}_0$ is to simply check which nodes from $\bb{Q}$ lie within $\mathcal{DEB}_0$. First, before checking whether \eqref{eq_dynamic_expanded_bubble} is satisfied for $\bb{q}_k$, i.e., if $\bb{q}_k \in \mathcal{DEB}_0$, it is necessary that $v_{obs} (t_k-t_0) \prec \bb{d}(t_0)$ (where ``$\prec$'' and ``$\succ$'' stand for element-wise comparison of each element from $\bb{d}$). Then, the sufficient condition \eqref{eq_dynamic_expanded_bubble} is checked. If satisfied, the spine $\overline{\bb{q}_0 \bb{q}_k}$ is collision-free for $t\in[t_0, t_k]$, i.e., $\overline{\bb{q}_0 \bb{q}_k}\in \mathcal{DEB}_0$, thus it is added to $DBur_0$ (e.g., spines $\overline{\bb{q}_0 \bb{q}_1},\dots, \overline{\bb{q}_0 \bb{q}_6}$ from Fig. \ref{fig_dynamic_bur}). Otherwise, the procedure terminates returning $DBur_0$, as well as in case $v_{obs} (t_k-t_0) \succeq \bb{d}(t_0)$ implying that all nodes $\bb{q}_m \notin \mathcal{DEB}_0$, $\forall m\in\{k,\dots,N\}$, (e.g., nodes $\bb{q}_7,\bb{q}_8,\bb{q}_9$ in Fig. \ref{fig_dynamic_bur}). Precisely, $i$-th robot's link at $\bb{q}_m$ would penetrate through $\mathcal{P}_{i,j}(t_m)$ regarding the closest $\mathcal{WO}_j$. Since we compute the signed values of $\bb{d}(t_m)$, it would result in $d_i(t_m) < 0$. 

Note that Alg. \ref{Pseudocode Dynamic bur} provides an approximate determination of the border of $\mathcal{DEB}_0$, i.e., $t^*\approx t_{k-1}$ (e.g., $t^*\approx t_6$ in Fig. \ref{fig_dynamic_bur}), when the robot moves along the specified trajectory $\bb{\pi}(t)$.

\vspace{-0.1cm}
\begin{definition}
	The set of collision-free spines $\overline{\bb{q}_0 \bb{q}_k}\in \mathcal{DEB}(\widetilde{\bb{q}}_0, \bb{d}(t_0), v_{obs})$, $k\in\{1,2,\dots,N\}$, represents the structure we refer to as dynamic bur.
\end{definition}
\vspace{-0.1cm}

\SetEndCharOfAlgoLine{}
\begin{algorithm}[t]
	\SetAlgorithmName{Algorithm}{}
	\Indp
	\caption{Dynamic generalized bur -- $\mathtt{computeDGBur}$}
	\label{Pseudocode Dynamic generalized bur}
	\small
	
	\KwIn{$\bb{\pi}(t) = \bb{\pi}[\bb{q}_0,\bb{q}_f]$, $\Delta t$, $\bb{d}(t_0)$, $v_{obs}$, $K$} 
	
	\KwOut{$DGBur$}
	
	$DGBur \gets \varnothing$ \\
	
	$k,\, chain\_size \gets 0$ \\
	
	\While{$\bb{q}_k \neq \bb{q}_f \,\And\, chain\_size < K$}
	{
		$DBur_k \gets \mathtt{computeDBur}(\bb{\pi}[\bb{q}_k, \bb{q}_f],\, \Delta t,\, \bb{d}(t_{k}),\, v_{obs})$ \label{Alg3 computeDBur} \\
		
		\If{$DBur_k \neq \varnothing$}
		{\label{Alg3 Q* empty check}
			
			$\overline{\bb{q}_k \bb{q}_m} \gets DBur_k(\textbf{end})$ \label{Alg3 q_m} \\
			
			$t_m \gets \mathtt{getTime}(\bb{q}_m)$ \label{Alg3 getTime} \\
			
			$\bb{\mathcal{P}}(t_m) \gets \mathtt{updatePlanes}(\bb{\mathcal{P}}(t_k),\, v_{obs}(t_m - t_k))$ \label{Alg3 updatePlanes} \\
			
			$\bb{d}(t_m) \gets \mathtt{getDistancesToPlanes}(\bb{q}_m,\, \bb{\mathcal{P}}(t_m))$ \label{Alg3 getDistancesToPlanes} \\
			
			$DGBur \gets [DGBur,\, DBur_k]$ \label{Alg3 update Q} \\
			
			$chain\_size \gets chain\_size + 1$ \label{Alg3 chain_size} \\
			
			$k \gets m$ \\
		}
		\Else
		{
			\Return
		}
	}
\end{algorithm}
\setlength{\textfloatsep}{0pt}

\vspace{-0.3cm}
\subsection{Chaining Dynamic Expanded Bubbles and Burs}
\label{Subsec. Chaining Dynamic Expanded Bubbles and Burs}
We are particularly interested in further exploration of $\mathcal{C}_{free}$ after reaching the border of $\mathcal{DEB}$ (e.g., $\bb{q}_6$ in Fig. \ref{fig_dynamic_bur}) since spines $\overline{\bb{q}_0\bb{q}_k} \notin DBur_0$ do not have to be in a collision (e.g., spines $\overline{\bb{q}_0 \bb{q}_7}$, $\overline{\bb{q}_0 \bb{q}_8}$ and $\overline{\bb{q}_0 \bb{q}_9}$ from Fig. \ref{fig_dynamic_bur}). To this end, we propose a simple and fast approach within Alg. \ref{Pseudocode Dynamic generalized bur} for computing a so-called \textit{dynamic generalized bur} $DGBur$. The proposed procedure involves the following steps:

\textbf{Step 1} (lines \ref{Alg3 computeDBur}--\ref{Alg3 Q* empty check}): First, $DBur_k$ is generated by the $\mathtt{computeDBur}$ function (provided by Alg. \ref{Pseudocode Dynamic bur}) using intermediate nodes from the trajectory $\bb{\pi}[\bb{q}_k, \bb{q}_f]$, $\forall k\in\{m,\dots,N-1\}$, for some $m\in\{0,1,\dots,N-1\}$, (e.g., yellow spines from $\bb{q}_0$ towards $\bb{q}_1, \bb{q}_2, \bb{q}_3 \in \mathcal{DEB}_0$, from $\bb{q}_3$ towards $\bb{q}_4,\dots,\bb{q}_{7} \in \mathcal{DEB}_3$, and from $\bb{q}_7$ towards $\bb{q}_8,\dots,\bb{q}_{11} \in \mathcal{DEB}_7$ in Fig. \ref{fig_dynamic_gbur} (a)). If $DBur_k$ exists, the subsequent steps can be executed. Otherwise, the procedure returns $DGBur$ computed so far.

\textbf{Step 2} (lines \ref{Alg3 q_m}--\ref{Alg3 getTime}): The last checked node $\bb{q}_m$, obtained from the corresponding spine $\overline{\bb{q}_k\bb{q}_m} \in DBur_k$, becomes the root for a new $\mathcal{DEB}_m$ (e.g., $\bb{q}_3$ and $\bb{q}_7$ become the root of $\mathcal{DEB}_3$ and $\mathcal{DEB}_7$ in Fig. \ref{fig_dynamic_gbur}\,(a), respectively). The routine $\mathtt{getTime}$ provides the time instance $t_m$ for $\bb{q}_m \in \bb{\pi}[\bb{q}_k, \bb{q}_f]$.

\textbf{Step 3} (line \ref{Alg3 updatePlanes}): To enable computing $\mathcal{DEB}_m$, the function $\mathtt{updatePlanes}$ provides new planes $\bb{\mathcal{P}}(t_m)$ by capturing the motions of all planes $\bb{\mathcal{P}}(t_k)$ (caused by the motion of obstacles) during the time $t\in[t_k, t_m]$. 

\textbf{Step 4} (line \ref{Alg3 getDistancesToPlanes}): Since the robot has moved from $\bb{q}_k$ to $\bb{q}_m$, new underestimates of $\bb{d}$ can be easily acquired (see \cite{lacevic2020gbur}) by the $\mathtt{getDistancesToPlanes}$ function as distances to the new planes $\bb{\mathcal{P}}(t_m)$, when the robot assumes $\bb{q}_m$ (e.g., $\bb{q}_3$ and $\bb{q}_7$ in Fig. \ref{fig_dynamic_gbur} (c, d)). Interestingly, the proposed way of computing distances may even increase the size of $\mathcal{DEB}_m$ in case the $i$-th link moves away from $\mathcal{WO}_{j}$ providing $d_i(t_m) > d_i(t_k)$.

\textbf{Step 5} (line \ref{Alg3 update Q}--\ref{Alg3 chain_size}): All spines from the newly computed bur $DBur_k$ are concatenated within $DGBur$. The size of a chain of connected $\mathcal{DEB}$s (it is the same as the number of generated corresponding dynamic burs) is tracked via the local variable $chain\_size$. It also denotes the number of extensions/layers of the dynamic generalized bur (e.g., Fig. \ref{fig_dynamic_gbur} (a) depicts 3 layers in total).

The logic from all steps 1--5 can be repeated for the new $k \gets m$ until all intermediate nodes from $\bb{\pi}[\bb{q}_0, \bb{q}_f]$ turn out to be collision-free (i.e., while $\bb{q}_k \neq \bb{q}_f$), or until the maximal number of layers $K$ is reached, or until $DBur_k=\varnothing$.

\vspace{-0.1cm}
\begin{definition}
	The set of collision-free spines $\overline{\bb{q}_k \bb{q}_i}\in \mathcal{DEB}(\widetilde{\bb{q}}_k, \bb{d}(t_k), v_{obs})$, $\forall i\in\{k+1,\dots,m\}$, for some $k\in\{0,1,\dots,$ $N-1\}$ and $m\in\{k+1,\dots,$ $N\}$, represents the structure we refer to as dynamic generalized bur.
\end{definition}
\vspace{-0.1cm}

\begin{figure}[t]
	\centering
	\includegraphics[width=\linewidth]{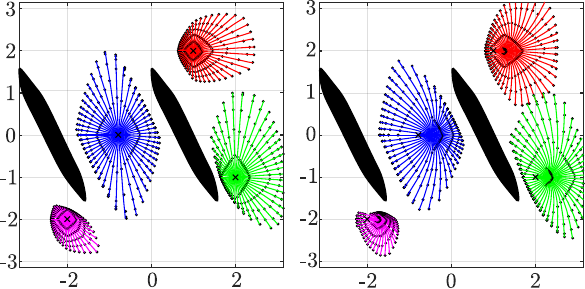}
	\caption{The set of collision-free trajectories in $\{q_1,q_2\}$-plane, generated using $DGBur$, from the configurations depicted in Fig. \ref{fig_dynamic_exp_bubbles_example} (left).}
	\label{fig_dynamic_gbur_example}
	\vspace{-0.05cm}
\end{figure}

Fig. \ref{fig_dynamic_gbur_example} shows the set of collision-free trajectories generated from the configurations depicted in Fig. \ref{fig_dynamic_exp_bubbles_example} (left), where Alg. \ref{Pseudocode Dynamic generalized bur} is applied for their collision checking using $K=5$ layers. We use $\Delta t = 100\,\mathrm{[\mu s]}$, and quintic polynomial as a local trajectory with $t_f = 1\,\mathrm{[s]}$, where $\|\bb{q}_f-\bb{q}_0\| = 2\,\mathrm{[rad]}$. In both subfigures, the initial acceleration for trajectory time parameterization is set to zero, while the initial velocity is zero in the left and $\left[2, 0\right]^T$ in the right subfigure.

\vspace{-0.3cm}
\subsection{Generating Trajectories}
\label{Subsec. Generating Trajectories}
Suppose that a local trajectory $\bb{\pi}_{curr}(t) \equiv \bb{\pi}[\bb{q}_{curr},\bb{q}_{next}]$ for $t\in[t_0,t_f]$ is generated in each algorithm iteration (Fig. \ref{fig_dynamic_gbur} (a)). For instance, a single quintic polynomial for each robot's joint renders our trajectory, and we refer to it as a \textit{current spline}, without the loss of generality.

\vspace{-0.1cm}
\begin{definition}
	The configuration $\bb{q}(t)$ is considered safe at the time instance $t$ if the robot can safely stop from $\bb{q}$ without colliding with the environment when the stopping starts at $t$.
\end{definition}
\vspace{-0.25cm}
\begin{definition}
	The spline $\bb{\pi}(t) = \bb{\pi}[\bb{q}_0, \bb{q}_f]$ is called safe, if it remains collision-free $\forall t\in[t_0, t_f]$. Otherwise, it is unsafe.
\end{definition}
\vspace{-0.1cm}

The process of generating splines is a part of the $\mathtt{updateCurrState}$ routine from Alg. \ref{Pseudocode DRGBT algorithm} depending on the parameter $safe\_on$ indicating whether the spline is guaranteedly safe for the environment. Our approach focuses on computing the spline with its shortest possible duration time $t_f$ while satisfying the set of constraints $\mathcal{K}$ (i.e., maximal joint velocity, acceleration, and jerk). The initial position $\bb{\pi}_{curr}(t_0) = \bb{q}_{curr}$, velocity $\dot{\bb{\pi}}_{curr}(t_0)$, and acceleration $\ddot{\bb{\pi}}_{curr}(t_0)$ are inherited from the spline computed in the previous iteration. The final position $\bb{\pi}_{curr}(t_f) = \bb{q}_{next}$ remains fixed. In case $safe\_on$ is false, the final velocity $\dot{\bb{\pi}}_{curr}(t_f)$ is estimated using two configurations, $\bb{q}_{curr}$ and $\bb{q}_{next}$, in order to achieve a minimal value for $t_f$ to account for $\mathcal{K}$. Otherwise, when $safe\_on$ is true, $\dot{\bb{\pi}}_{curr}(t_f) = \bb{0}$. For simplicity, the final acceleration is always set to zero, i.e., $\ddot{\bb{\pi}}_{curr}(t_f) = \bb{0}$. Using these six equations suffices to compute the quintic spline $\bb{\pi}_{curr}(t)$.

The spline $\bb{\pi}_{emg}(t) = \bb{\pi}[\bb{q}_{new},\bb{q}_{stop}]$ is called \textit{emergency spline} (Fig. \ref{fig_dynamic_gbur} (a)). The initial position, velocity, and acceleration are known from $\bb{\pi}_{curr}(t)$ as $\bb{\pi}_{emg}(0) = \bb{\pi}_{curr}(t_{new}) = \bb{q}_{new}$,  $\dot{\bb{\pi}}_{emg}(0) = \dot{\bb{\pi}}_{curr}(t_{new})$, and $\ddot{\bb{\pi}}_{emg}(0) = \ddot{\bb{\pi}}_{curr}(t_{new})$. The final position $\bb{\pi}_{emg}(t_{stop}') = \bb{q}_{stop}$, where $t_{stop}' = t_{stop}-t_{new}$, is temporarily unknown and will be consequently computed, while the final velocity and acceleration are always set to zero, i.e., $\dot{\bb{\pi}}_{emg}(t_{stop}') = \bb{0}$ and $\ddot{\bb{\pi}}_{emg}(t_{stop}') = \bb{0}$. Using these five equations suffices to compute the quartic spline $\bb{\pi}_{emg}(t)$. The open-source implementation is available \href{https://github.com/robotics-ETF/RPMPLv2/tree/main/src/planners/trajectory}{here}.


\vspace{-0.4cm}
\subsection{Application of Dynamic Expanded Bubbles and Burs}
\label{Subsec. Application of Dynamic Expanded Bubbles and Burs}
The aim is to establish whether the candidate spline $\bb{\pi}_{curr}(t)$ lies completely (or to what extent, if partially) within a single or chain of $\mathcal{DEB}$s. For the purpose of DRGBT algorithm, we require $\bb{\pi}_{curr}(t)$ (or its part) to be collision-free (safe) during the current algorithm iteration, as well as until the end of \textbf{T1} in the next iteration. Since new spline will be computed in the next iteration (it must happen anytime between $T$ and $T + e_1$ as a part of the $\mathtt{updateCurrState}$ routine from \textbf{T1} within Alg. \ref{Pseudocode DRGBT algorithm}), we need to ensure that the current spline is safe for $t\in[t_0, t_{new}]$, where $t_0=0$ and $t_{new} = T + e_1 < t_f$. Since the robot's velocity and acceleration are generally non-zero in $\bb{\pi}_{curr}(t_{new}) = \bb{q}_{new}$, a new safe spline $\bb{\pi}_{emg}(t)$ must be computed from $\bb{q}_{new}$ enabling the robot to stop at the configuration $\bb{q}_{stop}$. In special case when $T + e_1 \geq t_f$, then we choose $t_{new} = t_f$ and $\bb{q}_{stop} = \bb{q}_{next}$, thus the computation of $\bb{\pi}_{emg}(t)$ is not required. 

Now we describe the procedure for checking the safety of the composite spline $\bb{\pi}_{comp}(t) = \bb{\pi}[\bb{q}_{curr},\bb{q}_{stop}]$, which consists of two connected ``subsplines'', $\bb{\pi}[\bb{q}_{curr},\bb{q}_{new}]$ and $\bb{\pi}[\bb{q}_{new},\bb{q}_{stop}]$. One of the approaches would be to assign all its intermediate nodes as candidates for root configurations for generating $\mathcal{DEB}$s. If all bubbles exist ($\bb{d}(t_k) \succ \bb{0}$), they form a chain of $\mathcal{DEB}$s containing $\bb{\pi}_{comp}(t)$, thus the spline is considered safe. This approach enables the desired safety features, however, it requires a large number of $\mathcal{DEB}$s, and each of them further requires the computation of forward kinematics and corresponding enclosing radii $r_{i,j}$. Thus, we describe the alternative approach in the sequel.

The idea is to simply utilize Alg. \ref{Pseudocode Dynamic generalized bur} to generate $DGBur$ towards $\bb{\pi}_{comp}(t)$. Then, it is sufficient to check whether $\overline{\bb{q}_{curr}\bb{q}_{stop}} \in DGBur$. If yes, all generalized bur spines exist (e.g., yellow spines in Fig. \ref{fig_dynamic_gbur} (a)), i.e., $DGBur$ contains all intermediate nodes and their corresponding spines, implying $\bb{\pi}_{comp}(t)$ will remain safe for $t\in[t_0, t_{stop}]$. Otherwise, $\bb{\pi}_{comp}(t)$ may possibly be unsafe.

\begin{figure}[t]
	\centering
	\includegraphics[width=0.8\linewidth]{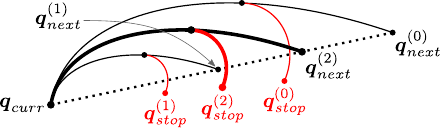}
	\caption{Using bisection method to find a safe spline.}
	\label{fig_splines_bisection_method}
\end{figure}

In case $\bb{\pi}_{comp}(t)$ is possibly unsafe, then $\bb{q}_{next}$ should be changed and a new spline $\bb{\pi}_{comp}(t)$ acquired. Finally, the safety of that spline can be checked. We propose using bisection method as shown in Fig. \ref{fig_splines_bisection_method} over the horizon spine $\overline{\bb{q}_{curr} \bb{q}_{next}^{_{(0)}}}$ (dotted black line) in order to find a new convenient node $\bb{q}_{next}$ that will yield a safe spline. For instance, assume $\bb{\pi}[\bb{q}_{curr},\bb{q}_{stop}^{(0)}]$ is unsafe. Then, a new next configuration is computed as $\bb{q}_{next}^{(1)} = (\bb{q}_{curr} + \bb{q}_{next}^{(0)})/2$, and $\bb{\pi}[\bb{q}_{curr},\bb{q}_{stop}^{(1)}]$ is checked for safety. If it is safe, the process might terminate. However, in search for a longer spline, a more distant node is computed as $\bb{q}_{next}^{(2)} = (\bb{q}_{next}^{(1)} + \bb{q}_{next}^{(0)})/2$, and $\bb{\pi}[\bb{q}_{curr},\bb{q}_{stop}^{(2)}]$ is checked for safety as illustrated by Fig. \ref{fig_dynamic_gbur} (a) in detail. The described problem can be formulated as: 
\vspace{-0.2cm}
\begin{problem}[Computing a safe spline]
	\label{Problem Computing a safe spline}
	For a given desired next configuration $\bb{q}_{next}$, we aim at computing the longest possible spline from the current configuration $\bb{q}_{curr}$ towards $\bb{q}_{next}$ which is guaranteedly safe, and which has minimal final time $t_f$ while respecting the set of constraints $\mathcal{K}$.
\end{problem}
\vspace{-0.2cm}
Such a bisection method can run until the maximal number of iterations exceeds. If this approach never finds a safe spline, there is always an ultimate solution -- immediately execute an emergency stopping from $\bb{q}_{curr}$, thus safely stopping the robot at the configuration $\bb{q}_{stop}'$ (e.g., dotted red line in Fig. \ref{fig_dynamic_gbur} (a)). The stopping trajectory $\bb{\pi}[\bb{q}_{curr}, \bb{q}_{stop}']$ will also be safe since it is a part of the safe spline that was computed in the previous algorithm iteration. 

One can notice that the proposed approach (which is based on Alg. \ref{Pseudocode Dynamic generalized bur}) relies on discrete collision checking of the spline depending on $\Delta t$. Nevertheless, the value of $\Delta t$ can be related to the sampling time of the robot's controller. On the other hand, the fact that the robot's links are approximated by bounding capsules (see Subsec. \ref{Subsec. Scenario Setup}) provides a larger safety margin that may compensate for the discrete approximation.

The following theorem encapsulates the safety features of the above-described approach utilizing Algs. \ref{Pseudocode Dynamic bur} and \ref{Pseudocode Dynamic generalized bur}. 

\vspace{-0.2cm}
\begin{theorem}[Guaranteed safe motion of the robot in dynamic environments under bounded obstacle velocity]
	\label{Theorem Guaranteed safe motion of the robot in dynamic environments under bounded obstacle velocity}
	If a configuration $\bb{q}_{new}$ is safe and the maximal obstacle velocity is limited to $v_{obs}$, the robot will follow the safe spline $\bb{\pi}[\bb{q}_{curr},\bb{q}_{new}]$ during the time $t\in[t_0, t_{new}]$, thus moving from the current configuration $\bb{q}_{curr}$ to a new configuration $\bb{q}_{new}$ satisfying the set of kinematic constraints $\mathcal{K}$. Otherwise, the robot will immediately execute stopping from $\bb{q}_{curr}$ to another safe configuration $\bb{q}_{stop}'$ by following the safe spline $\bb{\pi}[\bb{q}_{curr},\bb{q}_{stop}']$. Therefore, a solution satisfying $\mathcal{K}$ always exists leading the robot safely within the boundaries of the chain of connected dynamic expanded bubbles.
\end{theorem}

\vspace{-0.3cm}
\begin{corollary}
	The robot is always located at a \textit{safe configuration}. Therefore, it has arrived in $\bb{q}_0 = \bb{q}_{curr}$ only because it represents a safe configuration, meaning that it can stop safely as needed using the emergency spline $\bb{\pi}[\bb{q}_{curr},\bb{q}'_{stop}]$, which is needed if the new safe spline $\bb{\pi}[\bb{q}_{curr},\bb{q}_{stop}]$ cannot be computed. The theorem provides sufficient conditions, thus the possible violation of its conditions does not necessarily imply the occurrence of a collision.
\end{corollary}

\vspace{-0.3cm}
\begin{corollary}
	The robot will automatically scale its velocity depending on the size of obtained $\mathcal{DEB}$s that further depends on $d_{i}(t_k)$, as well as on the direction of the robot's links motion. For smaller values of $d_{i}(t_k)$, and particularly when the robot's $i$-th link approaches $\mathcal{WO}_j$, $\mathcal{DEB}$s become smaller. Consequently, corresponding safe splines get shorter. Regarding Problem \ref{Problem Computing a safe spline} as well as $\dot{\bb{q}}_{next} = \bb{0}$ and $\ddot{\bb{q}}_{next} = \bb{0}$, the obtained maximal velocity on $\bb{\pi}[\bb{q}_{curr},\bb{q}_{next}]$ may decrease if $\bb{q}_{next}$ is closer to $\bb{q}_{curr}$ (e.g., the obtained maximal velocity on $\bb{\pi}[\bb{q}_{curr},\bb{q}_{next}^{_{(2)}}]$ is smaller or equal to the one on $\bb{\pi}[\bb{q}_{curr},\bb{q}_{next}^{_{(0)}}]$ when considering Fig. \ref{fig_splines_bisection_method}).
\end{corollary}

\vspace{-0.5cm}
\section{Simulation Study}
\label{Sec. Simulation Study}
This section provides an extensive simulation study\footnote{The implementation of DRGBT algorithm in C++ is available \href{https://github.com/robotics-ETF/RPMPLv2/tree/main/src/planners/drbt}{here}. We use The Kinematics and Dynamics Library (\href{https://github.com/orocos/orocos_kinematics_dynamics}{KDL}), Flexible Collision Library (\href{https://github.com/flexible-collision-library/fcl}{FCL}) \cite{pan2012fcl}, and \href{https://github.com/jlblancoc/nanoflann}{nanoflann library} \cite{blanco2014nanoflann}. It is worth mentioning that using FCL is optional since we implemented our \href{https://github.com/robotics-ETF/RPMPLv2/blob/main/src/state_spaces/real_vector_space/CollisionAndDistance.cpp}{library} which is capable of fast handling simple primitive objects such as boxes, spheres and capsules. The simulation was performed using the laptop PC with Intel\textregistered\, Core\texttrademark\, i7-9750H CPU @ 2.60 GHz $\times$ 12 with 16 GB of RAM, with the code compiled to run on a single core of the CPU without any GPU acceleration.} with two distinct goals. The first goal is to subject DRGBT to a large number of scenarios to establish the appropriate deadline allocation for a real-time scheduling algorithm. Another goal is to utilize such scenarios to compare DRGBT to state-of-the-art methods within a randomized trial through Sec. \ref{Sec. Comparison to State-of-the-art Methods}.


\vspace{-0.3cm}
\subsection{Scenario Setup}
\label{Subsec. Scenario Setup}
The simulation study deals with 15 scenario types, with each type assuming a given number of obstacles from the set $\{0, 1, 2, \dots, 10, 20, 30, 40, 50\}$. Within each type, the model of the UFactory xArm6 robot is exposed to 1000 different simulation runs (resulting in a total of 15000 runs) with randomly generated circumstances, as described in the sequel.

\begin{figure}[t]
	\centering
	\includegraphics[width=0.8\linewidth]{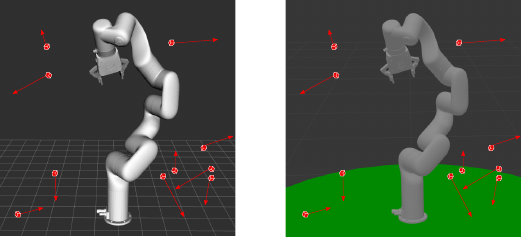}
	\caption{An example of the initial configuration alongside 10 random obstacles in RViz (left) and Gazebo (right)}
	\label{fig_RViz_Gazebo_xarm6_10obs}
	\vspace{-0.05cm}
\end{figure}

\begin{figure*}[t]
	\centering
	\includegraphics[width=\linewidth]{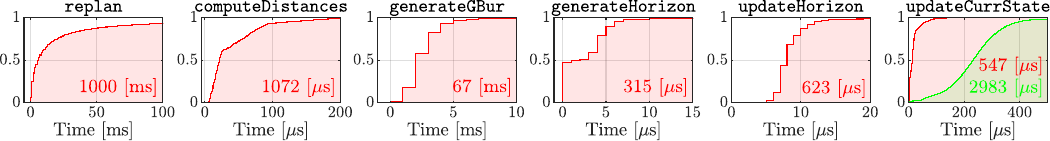}
	\vspace{-0.6cm}
	\caption{Corresponding CDFs for the routines: $\mathtt{replan}$, $\mathtt{computeDistances}$, $\mathtt{generateGBur}$, $\mathtt{generateHorizon}$, $\mathtt{updateHorizon}$ and $\mathtt{updateCurrState}$. Execution times are indicated in each subfigure. Legend: DRGBT (red) and DRGBT-safe (green).}
	\label{fig_routines_CDF}
	\vspace{-0.4cm}
\end{figure*}
\begin{figure*}[t]
	\centering
	\includegraphics[width=\linewidth]{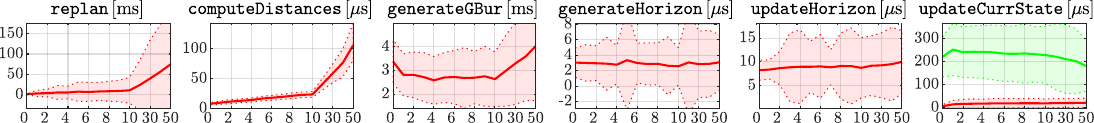}
	\vspace{-0.5cm}
	\caption{Mean response time (solid line) and standard deviation (dotted line) w.r.t. $N_{obs}$ for the routines: $\mathtt{replan}$, $\mathtt{computeDistances}$, $\mathtt{generateGBur}$, $\mathtt{generateHorizon}$, $\mathtt{updateHorizon}$ and $\mathtt{updateCurrState}$. Legend: DRGBT (red) and DRGBT-safe (green).}
	\label{fig_routines}
	\vspace{-0.65cm}
\end{figure*}

\textbf{1)} Obstacles are placed w.r.t. uniform distribution within $\mathcal{W}$. For the sake of simplicity, we assume that each obstacle is an axis-aligned bounding box (AABB) with fixed dimensions of $1\times 1\times 1\,\mathrm{[cm^3]}$. Each obstacle is assigned a random constant velocity, where its magnitude is limited to\footnote{We choose such value according to the ISO TS 15055 recommendations, which specify that in human-robot collaboration scenarios, if the human’s velocity is not being monitored, the system should assume a velocity of $1.6\,\mathrm{[\frac{m}{s}]}$ in the direction that minimizes the human-robot separation distance.} $1.6\,\mathrm{[\frac{m}{s}]}$. The choice of these relatively small dimensions stems from the motivation to increase the likelihood that the algorithm finds a solution in case of many obstacles (e.g., 50). The primary objective of this study is to evaluate how the algorithm scales with the number of obstacles processed. A large obstacle count may also arise from decomposing a few relatively larger obstacles into multiple geometric primitives (e.g., AABBs). Apparently, larger obstacles cause frequent obstructions of goal configuration and thus decimating the success rates of tested algorithms. Conversely, scenarios with small obstacles are not necessarily trivial, especially when dealing with a large number of them. Since the robot occupies significant volume (particularly when approximated by bounding volumes, e.g., capsules), even fast moving point-like obstacles pose a substantial threat for collision.

\textbf{2)} The used workspace $\mathcal{W}$ is bounded by a sphere with the radius of $1.5\,\mathrm{[m]}$ and the center in $(0, 0, 0.267)\,\mathrm{[m]}$, since this point represents the robot's base top point. If an obstacle reaches the limit of $\mathcal{W}$ or gets closer to $\max\{\frac{v_{obs}}{\omega_{max,1}}, r_1\}$ from the robot's base with a radius $r_1$, it just bounces back to $\mathcal{W}$ respecting the law of reflection while maintaining the velocity magnitude $v_{obs}$. In other words, random obstacles cannot exist in close vicinity of the manipulator's base to prevent the inevitable collisions.

\textbf{3)} The configurations $\bb{q}_{start}$ and $\bb{q}_{goal}$ are generated randomly in each run, with the condition $\rho(\bb{q}_{start}, \bb{q}_{goal}) > \rho_0$, with $\rho$ being a $\mathcal{C}$-space metric from \cite{kavraki1996probabilistic}, and $\rho_0$ a proper threshold. Fig. \ref{fig_RViz_Gazebo_xarm6_10obs} depicts an example of $\bb{q}_{start}$ alongside 10 random obstacles (red boxes) and their velocity vectors illustrated in RViz (left) and Gazebo (right). The manipulator is mounted on the round green table with the radius of $67\,\mathrm{[cm]}$ as shown in Gazebo, which is also considered a static obstacle. 

\textbf{4)} To facilitate collision/distance checks, the robot's links are approximated by bounding capsules. Moreover, self-collision checking is implemented and taken into account concurrently during the (dynamic) generalized burs computation. 

\textbf{5)} To construct a smooth trajectory from $\bb{q}_{curr}$ to $\bb{q}_{next}$ (within $\mathtt{updateCurrState}$ from Alg. \ref{Pseudocode DRGBT algorithm}), we use quintic splines, since the real xArm6 manipulator has to meet constraints $\mathcal{K}$ on maximal joint velocity, acceleration and jerk. 

\textbf{6)} The initial parameters used by DRGBT are: 
\vspace{-0.05cm}
\begin{itemize}[leftmargin=0.35cm]
	\item when using RGBMT* \cite{covic2023asymptotically} as a replanner, the terminating condition is set to finding the first feasible path;
	\item the maximal velocity, acceleration, and jerk of each robot's joint are $\bb{\omega}_{max} = \bb{\pi}_{n\times 1}\,\mathrm{[\frac{rad}{s}]}$, $\bb{\alpha}_{max} = \bb{20}_{n\times 1}\,\mathrm{[\frac{rad}{s^2}]}$ and $\bb{j}_{max} = \bb{500}_{n\times 1}\,\mathrm{[\frac{rad}{s^3}]}$, respectively, taken from the datasheet of xArm6 robot;
	\item the initial horizon size is $N_{h_0}=10$;
	\item the critical workspace distance is $d_{crit} = 0.05\,[\mathrm{m}]$;
	\item the thresholds for assessing whether the replanning is required are $w_{min} = \overline{w}_{min} = 0.5$;
	\item the maximal number of attempts when modifying bad or critical nodes is $10$.
\end{itemize}

In theory, any planning algorithm can be used for replanning within DRGBT. In \cite{covic2021path}, the RGBT-Connect \cite{lacevic2020gbur} was used for its quickness. In this paper, we also investigate the usage of RGBMT* \cite{covic2023asymptotically} due to its asymptotic optimality feature. It has shown to be quick during replanning phases. Generally, when using any asymptotically optimal planner for replanning, its default terminating condition should be \textit{when the first feasible path is found}. Otherwise, the user might specify the exact replanning time, enabling the resulting path to be possibly improved until the available replanning time expires. More precisely, whenever \textbf{T2} completes before its relative deadline, the processor becomes idle, as in Fig. \ref{fig_FPS_scheduling}. In this case, the replanning may be queried more often, i.e., until the time $D_2$ exceeds. Since more predefined paths may be obtained, one of them should be selected according to some criteria, such as path length or path safety. Unfortunately, the evaluation of these criteria is often time-consuming, which makes them inconvenient for real-time motion generation. 

\vspace{-0.35cm}
\subsection{Simulation without Time Constraints}
\label{Subsec. Simulation without Time Constraints}
\vspace{-0.05cm}
This part of the study aims at collecting the data about the response and execution times of each task, \textbf{T1} and \textbf{T2}, when no time constraints are imposed. Only the maximal runtime of DRGBT is set to $10\,\mathrm{[s]}$ and the maximal replanning time to $1\,\mathrm{[s]}$. The idea is to use the collected data to infer the proper choice of time constraints to be imposed on critical tasks within the real-time implementation. We aim to reveal how the algorithm success rate depends on these constraints. As it turns out, there is a ``sweet spot'' for the iteration time $T$ w.r.t. the resulting success rate. Clearly, such setup allows for the time-varying algorithm iterations, which calls for expressing the obstacle velocity in $\mathrm{[m/iteration]}$ instead of $\mathrm{[\frac{m}{s}]}$. We set the upper bound of velocity to $10\,\mathrm{[cm/iteration]}$. For instance, if the iteration is limited to $62.5\,\mathrm{[ms]}$, the maximal obstacle velocity becomes exactly $1.6\,\mathrm{[\frac{m}{s}]}$. The motivation behind the choice of these specific values will be revealed later. 

To simplify the time analysis of each task, we decompose it into routines that are analyzed separately. More precisely, each line from Alg. \ref{Pseudocode DRGBT algorithm} represents a single routine (a part of a task), which is characterized by its execution time. The black lines indicate that the corresponding routine consumes negligible time (less than $1\,\mathrm{[\mu s]}$). The blue and red lines require substantial time and are subjected to the following analysis. 

Fig. \ref{fig_routines_CDF} shows corresponding cumulative density functions (CDFs) for the routines $\mathtt{replan}$, $\mathtt{computeDistances}$, $\mathtt{generateGBur}$, $\mathtt{generateHorizon}$, $\mathtt{updateHorizon}$ and $\mathtt{updateCurrState}$. It reveals the completion success of each routine within the desired time. Maximum times are also indicated since graphics' views are limited to the region of interest. Moreover, mean response time and standard deviation w.r.t. $N_{obs}$ for the same routines are provided by Fig. \ref{fig_routines}. The procedures $\mathtt{generateHorizon}$, $\mathtt{updateHorizon}$ and $\mathtt{updateCurrState}$ do not seem to be scenario-dependent (depicted by blue lines in Alg. \ref{Pseudocode DRGBT algorithm}) while other routines are (depicted by red lines in Alg. \ref{Pseudocode DRGBT algorithm}). The response time is approximately linear w.r.t. $N_{obs}$. Note that the abscissa scale is not linear since the considered numbers of obstacles in separate scenarios do not form the arithmetic sequence. It is worth stressing that Figs. \ref{fig_routines_CDF} and \ref{fig_routines} show data for DRGBT (-safe) in red (green). Since the only difference appears in the $\mathtt{updateCurrState}$ routine, it is clearly indicated.

The slope coefficient of the approximated line for the $\mathtt{replan}$ procedure is clearly conditioned to the complexity of the used scenario. Therefore, the discussion about this routine needs to be elaborated in a more general way (Subsec. \ref{Subsec. Imposing Time Constraints}). As for the standard deviation, it is expected to be relatively large, since the replanning time depends on the ``distance'' from $\bb{q}_{curr}$ to $\bb{q}_{goal}$, among other things. In other words, faster replanning is expected when the robot comes closer to the goal. 


Generally, the worst-case scenario for the task occurs when its actual response time is equal to its execution/maximum time. In this regard, Fig. \ref{fig_routines_CDF} reveals that maximum time of \textbf{T1} is $e_1 \approx 70\,\mathrm{[ms]}$ for DRGBT, and $e_1 \approx 73\,\mathrm{[ms]}$ for DRGBT-safe,
at least for the considered scenarios. In extreme cases involving more than 50 obstacles, $e_1$ will likely become larger. To summarize, the most time-consuming routines are $\mathtt{replan}$ and $\mathtt{generateGBur}$, which calls for their dedicated treatment.

\vspace{-0.4cm}
\subsection{Imposing Time Constraints}
\label{Subsec. Imposing Time Constraints}
\vspace{-0.05cm}
First, we discuss the possible consequences in case the scheduler interrupts a task (or some of its routines), and devise the strategy of imposing time constraints, since, for instance, we cannot wait infinitely for replanning to complete. After determining time constraints, simulations are performed for each iteration time from the set $T\in\{10,20,\dots,120\}\,\mathrm{[ms]}$ using 15000 algorithm runs (1000 runs of each of 15 scenario types as described in Subsec. \ref{Subsec. Scenario Setup}).

If $\mathtt{generateHorizon}$ is interrupted, the horizon might contain fewer nodes than expected (e.g., some nodes from the predefined path may not be added to the horizon). Similarly, when $\mathtt{updateHorizon}$ is interrupted, the horizon will not be fully updated (e.g., lateral spines may not be generated, or some nodes and their properties from the previous iteration may not be fully updated). Next, the robot may remain at the same configuration if $\mathtt{updateCurrState}$ does not completely execute, which may be a severe problem. Clearly, such interruptions might corrupt the algorithm flow triggering undesirable behavior (e.g., the code may crash due to unpredictable errors in case exceptions are not properly handled by catch blocks), and possibly lead to the collision with obstacles.

On the other hand, if the scheduler interrupts $\mathtt{generate}$-$\mathtt{GBur}$, the only thing that happens is that the generalized burs ends up with fewer spines (at least one). This does not pose any problem since the computation of all spines from the generalized bur is not required to be completed, and the algorithm is still capable of proceeding to subsequent routines.
Horizon nodes, for which there is no time for the corresponding generalized spine to be computed, are simply omitted from the horizon. The process of removing nodes does not disturb any of the subsequent routines expressed by lines \ref{Alg1 computeNodeWeights}--\ref{Alg1 updateCurrState} from Alg. \ref{Pseudocode DRGBT algorithm}. It may, however, affect the algorithm's performance (success rate, algorithm time, path length, etc.).

Comparing $\mathtt{generateGBur}$ and $\mathtt{replan}$ procedures, the first one deserves higher priority, since the smaller horizon (with fewer spines) increases the collision probability, especially in cases when $d_c$ gets smaller. Clearly, this can be explained by the fact that $\mathtt{generateGBur}$ subsumes some of the APF features that locally control the robot, rendering this routine responsible for collision avoidance. On the other hand, should the collision occur, $\mathtt{replan}$ is not responsible, since it provides only a global picture to the robot, i.e., replanning just provides waypoints that should navigate the robot to the goal. This is the main reason why \textbf{T1} is assigned higher priority than \textbf{T2}.

In order to achieve the best performance of the algorithm, we adopt the following. The response time of \textbf{T1} is mostly determined by $\mathtt{generateGBur}$, thus its execution time will be limited, while all other routines from \textbf{T1} are allowed to take all the time they require, i.e., they will not be interrupted by the scheduler. \textbf{T2} may start, depending on whether it is triggered, immediately after \textbf{T1} completion.

\begin{figure}[t]
	\centering
	\includegraphics[width=\linewidth]{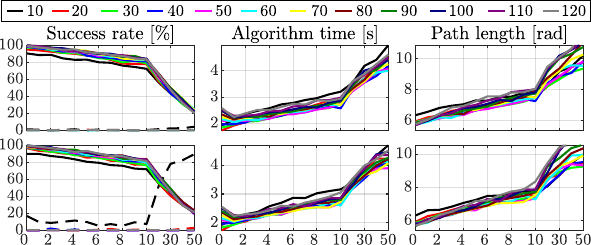}
	\vspace{-0.5cm}
	\caption{\textbf{Criteria} w.r.t. $N_{obs}$ for different values of $T$ (shown in legend in $\mathrm{[ms]}$) in cases $u_1 = 1$ (top) and $u_1 = 0.5$ (bottom).}
	\label{fig_DRGBT_results}
	\vspace{-0.3cm}
\end{figure}


\begin{figure}[t]
	\centering
	\includegraphics[width=\linewidth]{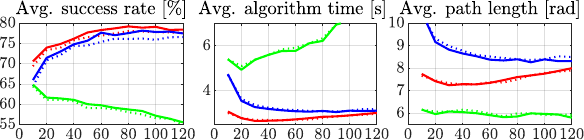}
	\vspace{-0.55cm}
	\caption{Averaged results of \textbf{criteria} w.r.t. $T\,\mathrm{[ms]}$ for {\color{red}DRGBT$^{(1)}$}, {\color{blue}DRGBT$^{(2)}$} and {\color{green}DRGBT-safe} in case $u_1\in\{1, 0.5\}$ (solid/dotted line)}
	\label{fig_DRGBT_final_results}
\end{figure}

\vspace{-0.3cm}
\subsection{Simulation with Imposed Time Constraints}
\label{Subsec. Simulation with Imposed Time Constraints}
Before discussing the obtained results, we define some criteria used for assessing the performance of DRGBT. \textit{Algorithm time} is defined as a product of the number of algorithm iterations and the iteration time $T$. More precisely, it determines the required time for the robot to reach the goal configuration from the start without colliding with obstacles. Such motion is marked as a \textit{successful} one, and we assign Euclidean-distance-based \textit{path length} in $\mathcal{C}$-space to the traversed path.

Fig. \ref{fig_DRGBT_results} shows how decreasing $T$ (period of \textbf{T1}) from $120\,\mathrm{[ms]}$ towards $10\,\mathrm{[ms]}$ affects different criteria used for assessing the performance of DRGBT, such as success rate, algorithm time and path length (denoted as \textbf{criteria} in the sequel) w.r.t. $N_{obs}$. Three top figures provide results in case when the utility of \textbf{T1} can go up to $u_1 = 1$, i.e., it may consume all the available time from the processor implying $e_1 = T$. Similar results are obtained by three bottom figures, except the utility of \textbf{T1} can go up to $u_1 = 0.5$ meaning $e_1 = 0.5T$. In both cases, \textbf{T2} may be executed as needed during the remaining time after \textbf{T1} completion respecting \eqref{eq_time_execution_task2}. Clearly, when $e_1 < 70\,\mathrm{[ms]}$ in Fig. \ref{fig_DRGBT_results}, the scheduler may interrupt \textbf{T1} execution, which makes it interesting to inspect potential consequences. It is worth mentioning that the success rate for zero random obstacles can be less than $100\,[\%]$, since the maximal available runtime of DRGBT can be exceeded.

Averaged results for \textbf{criteria} over all scenarios w.r.t. $T$ are shown in Fig. \ref{fig_DRGBT_final_results} for DRGBT$^{(1)}$ (using RGBMT* as replanner) and DRGBT$^{(2)}$ (using RGBT-Connect as replanner, where the same simulation study is carried out as for DRGBT$^{(1)}$). 
Clearly, decreasing $T$ provides more timely reactions to the changing environment, which yields a higher success rate. On the other hand, we cannot decrease it to zero, since it would cause the decrease of $e_1$ at the same time, which may interrupt \textbf{T1} more frequently. Thus, we identify the ``sweet spot'', which lies around $80\,\mathrm{[ms]}$ for DRGBT$^{(1)}$, and around $90\,\mathrm{[ms]}$ for DRGBT$^{(2)}$. As for the averaged time and path length of DRGBT$^{(1)}$, ``sweet spot'' can be found around $30\,\mathrm{[ms]}$. CDFs from Fig. \ref{fig_routines_CDF} reveal that $\mathtt{generateGBur}$ can be successfully completed in $99.9859\,[\%]$ cases within $30\,\mathrm{[ms]}$, while all other routines (except $\mathtt{replan}$) complete in $100\,[\%]$ cases. On the other hand, the best averaged time and path length of DRGBT$^{(2)}$ lies around $90\,\mathrm{[ms]}$.

\begin{figure}[t]
	\centering
	\includegraphics[width=\linewidth]{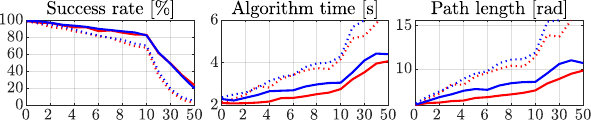}
	\vspace{-0.55cm}
	\caption{\textbf{Criteria} w.r.t. $N_{obs}$ for {\color{red}DRGBT$^{(1)}$} and {\color{blue}DRGBT$^{(2)}$} with and without replanning procedure (solid and dotted line, respectively).}
	\label{fig_results_without_replanning}
\end{figure}

Fig. \ref{fig_DRGBT_final_results} reveals that there is a tendency of worsening all criteria by reducing $T$ below its corresponding sweet spot. This can be supported by the fact that a shorter replanning time reduces the chance of successfully completing \textbf{T2}, as CDF in Fig. \ref{fig_routines_CDF} indicates for the $\mathtt{replan}$ routine. Therefore, the robot will only explore the surroundings, since the predefined path is mainly responsible for leading the robot to the goal. If we completely disable the replanning procedure in case of DRGBT$^{(1)}$ for $T = 50\,\mathrm{[ms]}$, and in case of DRGBT$^{(2)}$ for $T = 90\,\mathrm{[ms]}$, where $u_1 = 1$ in both cases, we obtain the results as in Fig. \ref{fig_results_without_replanning}. Thus, a global component of replanning enhances overall performance, yet it does not need to be mandatory since the robot may exploit only local information from the horizon (e.g., act as a reactive planner) and still find the goal.

On the other hand, if we disable \textbf{T1}, the goal will never be found since the procedure for updating $\bb{q}_{curr}$ would not exist. Dashed lines in two left figures from Fig. \ref{fig_DRGBT_results} indicate to what extent interrupting \textbf{T1} is responsible for collision between the robot and obstacles. Evidently, decreasing $e_1$ below $10\,\mathrm{[ms]}$ is primarily responsible for an increased likelihood of collisions. Obtained results suggest that it is desirable to assign enough processor time to \textbf{T1} (e.g., compare dashed black lines for $u_1 = 1$ and $u_1 = 0.5$). This further supports the choice of setting the higher priority for \textbf{T1} versus \textbf{T2}.

Accounting for all \textbf{criteria} previously discussed, the convenient choice for the DRGBT$^{(1)}$ period is $T^* = 50\,\mathrm{[ms]}$, implying the \textit{algorithm hard real-time frequency} $f_{alg}^* = 20\,\mathrm{[Hz]}$. This choice turns out as convenient for real-world applications as will be shown in Sec. \ref{Sec. Experiments}. The choice of $f_{alg}^*$ is also tightly related to typical frame rates ($\sim\,25\,\mathrm{[Hz]}$) of commercially available depth cameras. Unequivocally, the obtained results for DRGBT$^{(2)}$ suggest using $T^* = 90\,\mathrm{[ms]}$ yielding a relatively low frequency of $f_{alg}^* = 11.1\,\mathrm{[Hz]}$. Overall results from Fig. \ref{fig_DRGBT_final_results} suggest that using RGBMT* method in the replanning procedure would yield better performance than with RGBT-Connect, particularly in terms of the algorithm frequency.

Averaged results for DRGBT-safe according to \textbf{criteria} w.r.t. $T$ are also depicted in Fig. \ref{fig_DRGBT_final_results}. Clearly, the success rate is lower as expected, and decreases with increasing $T$. Nevertheless, it is worth reminding that all occurred collisions are type II collisions. In cca $\mathrm{2.94\,[\%]}$ cases, the maximal runtime is exceeded and those runs are labeled unsuccessful. Average algorithm time has significantly increased as expected since DRGBT-safe typically generates shorter trajectories/splines in order to ensure their existence within dynamic expanded bubbles. Consequently, a price to pay is a slower motion of the manipulator which may trigger collisions more frequently. Interestingly, average path length has improved, since adhering to the predefined path is usually the safest motion, which means the robot possibly does not explore potentially unsafe regions. To summarize, according to all \textbf{criteria}, DRGBT-safe performs better for smaller values of $T$ ($10$ to $20\,\mathrm{[ms]}$) implying higher algorithm frequencies $f_{alg}^* \in [50,100]\,\mathrm{[Hz]}$).

\vspace{-0.1cm}
\section{Comparison to State-of-the-art Methods}
\label{Sec. Comparison to State-of-the-art Methods}
In \cite{covic2021path}, the preliminary version of DRGBT was shown to outperform RRT$^\mathrm{X}$ algorithm \cite{otte2016rrtx} within a variety of scenarios. For the sake of completeness, we include RRT$^\mathrm{X}$ in the comparative study\footnote{To this end, we have implemented a C++ version of the algorithm available \href{https://github.com/robotics-ETF/RPMPLv2/blob/main/src/planners/rrtx/RRTx.cpp}{here}, by carefully following the original Julia code made available by the author \href{http://ottelab.com/html_stuff/code.html}{here}. Since the code is entirely reimplemented, we equipped it with the same trajectory parametrization as DRGBT in order to make sure that all kinematic constraints are satisfied.}. Moreover, we compare DRGBT to one of the recently proposed state-of-the-art algorithms. As the most representative example, we choose MARS algorithm \cite{tonola2023anytime}, primarily due to its proven performance in comparison with other competing algorithms (such as DRRT \cite{ferguson2006replanning}, Anytime DRRT \cite{ferguson2007anytime}, and MPRRT \cite{sun2015high}), and the availability of its C++ \href{https://github.com/JRL-CARI-CNR-UNIBS/OpenMORE}{implementation}. Each method uses its own default parameters.

\vspace{-0.2cm}
\subsection{Randomized trial}
\begin{figure}[t]
	\centering
	\includegraphics[width=\linewidth]{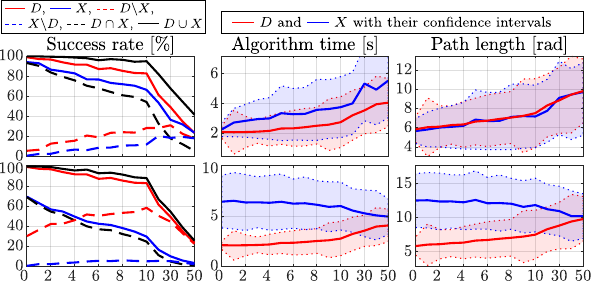}
	\vspace{-0.5cm}
	\caption{\textbf{Criteria} for {\color{red} DRGBT} ($D$) and {\color{blue} MARS} ($X = M$) -- top, and {\color{blue} RRT$^\mathrm{X}$} ($X = R$) -- bottom, w.r.t. $N_{obs}$.
	}
	\label{fig_DRGBT_random_scenarios_comparison}
\end{figure}

This test case represents the randomized trial, the same one used in Subsec. \ref{Subsec. Scenario Setup}. The results obtained for comparison DRGBT versus MARS and RRT$^\mathrm{X}$, according to \textbf{criteria}, are shown in Fig. \ref{fig_DRGBT_random_scenarios_comparison}. Additionally, the successful runs are categorized as follows: those where both methods successfully find a solution ($D\cap X$), those where either method finds a solution ($D\cup X$), and those where only one algorithm succeeds ($D\backslash X$ and $X\backslash D$), where $D$ (or $X\in\{M,R\}$) stands for a set of successful runs of DRGBT (or MARS / RRT$^\mathrm{X}$) w.r.t. $N_{obs}$. For scenarios with up to 10 obstacles, DRGBT demonstrates clear advantages for up to $20\,[\%]$ versus MARS, and for up to $50\,[\%]$ versus RRT$^\mathrm{X}$. For a higher $N_{obs}$, these advantages settle to approximately $10\,[\%]$ and $40\,[\%]$. A significant number of runs can be observed from $D\backslash M$ and $D\backslash R$. Another positive aspect of DRGBT is that $D$ closely aligns with $D\cup R$, and $R$ closely aligns with $D\cap R$.

Regarding algorithm time, DRGBT consistently finds solutions faster. However, the traversed path is approximately the same for DRGBT and MARS, while RRT$^\mathrm{X}$ yields significantly longer paths. Clearly, DRGBT outperforms the competing algorithms w.r.t. both criteria, proving its effectiveness in relatively difficult scenarios.

Overall, DRGBT demonstrates superior performance. Please note that, unlike DRGBT, MARS relies on the following:
\vspace{-0.05cm}
\begin{itemize}[leftmargin=0.35cm]
	\item Replanning is executed in parallel (at the frequency of $5\,\mathrm{[Hz]}$) with all other tasks of the algorithm;
	
	\item While searching for the initial paths, obstacles remain stationary. The algorithm consumes considerable time ($0.53\pm 0.68\,\mathrm{[s]}$) upfront to generate several initial paths;
	
	\item Kinematic constraints $\mathcal{K}$ are not explicitly considered during trajectory design, leading to violations of the maximal joint velocity in $10.5\,[\%]$ runs (with the values of $3.45 \pm 0.22\,\mathrm{[\frac{rad}{s}]}$), and the maximal joint acceleration in $34.2\,[\%]$ runs (with the values of $177.9 \pm 140.5\,\mathrm{[\frac{rad}{s^2}]}$).
\end{itemize}

In this context, the success rate of MARS can be regarded as ``conditional''. We presume that MARS algorithm can be modified to account for kinematic constraints. It remains an open question whether such modification would result in performance deterioration. Nevertheless, we refrained from intervening on the original code. On the other hand, if we relax the acceleration and jerk constraints for DRGBT, the obtained results become substantially better (e.g., success rate increases from $25\,\mathrm{[\%]}$ to $60\,\mathrm{[\%]}$ for $N_{obs} = 50$).

\vspace{-0.3cm}
\subsection{Scenarios with larger obstacles and narrow passages}
\begin{figure}[t]
	\centering
	\includegraphics[width=\linewidth]{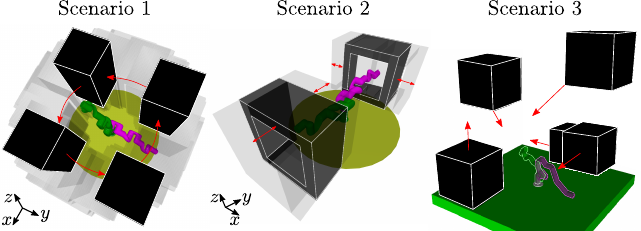}
	\caption{Three additional \href{https://www.youtube.com/watch?v=lG7q0PuFhG0}{scenarios} used in the comparison study.}
	\label{fig_DRGBT_additional_scenarios_comparison}
	\vspace{-0.1cm}
\end{figure}

In order to better evaluate the algorithm's performance, two more complex scenarios including larger obstacles and narrow passages, as shown in Fig. \ref{fig_DRGBT_additional_scenarios_comparison}, are added within the simulation comparison study. The scenarios are based on those from \cite{covic2021path}, this time set up consistently with the real-time implementation. The robot workspace is shown, where the initial and goal configuration are indicated (magenta and green, respectively). The obstacles (black) and their paths are also clearly indicated (gray with red arrows). In the first scenario, four prismatic obstacles follow a circular path in a horizontal plane. To reach the goal, the robot needs to squeeze through two moving adjacent obstacles. In the second scenario, the obstacles are two deformable tunnels. The first tunnel is periodically stretching and shrinking, while the second one is moving left and right along the $x$-axis. The robot end-effector needs to escape one tunnel and squeeze into another one. In both scenarios, obstacles' linear velocity is picked randomly from the range $(0, 0.3]\,\mathrm{[\frac{m}{s}]}$ for each of 1000 algorithm's runs. The maximal runtime is limited to $10\,\mathrm{[s]}$ for Scenario 1 and to $30\,\mathrm{[s]}$ for Scenario 2.

Tab. \ref{tab_DRGBT_additional_scenarios_results} shows the obtained results from the carried out study, where the success rate $\eta$, mean algorithm time $T_{alg}$, and mean path length $D$ are considered. Clearly, DRGBT outperforms both methods according to success rate, despite MARS violating kinematic constraints (e.g., on maximal joint velocity in $81.4\,\mathrm{[\%]}$ and $67.8\,\mathrm{[\%]}$ runs in scenarios 1 and 2, respectively; and on maximal joint acceleration and jerk in $100\,\mathrm{[\%]}$ runs in both scenarios). Scenario 2 is particularly difficult since the robot requires significant amount of time to escape one moving tunnel and to enter into another one. Tab. \ref{tab_DRGBT_additional_scenarios_results} proves that a trade-off between $\eta$, and $T_{alg}$ and $D$ exists. It is worth indicating that success rates of RRT$^\mathrm{X}$ are now worse than those in \cite{covic2021path} likely due to limited runtime.

\begin{table}[t]
	\caption{Results for Scenarios 1 and 2 from the comparison study.}
	\vspace{-0.2cm}
	\centering
	\resizebox{0.495\textwidth}{!}{%
		\begin{tabular}{@{}lccc|ccc@{}}
	\cmidrule(l){2-7}
	& \multicolumn{3}{c|}{Scenario 1} & \multicolumn{3}{c}{Scenario 2} \\ 
	\cmidrule(l){2-7} 
	& DRGBT     & MARS & RRT$^\mathrm{X}$  & DRGBT     & MARS  & RRT$^\mathrm{X}$  \\ \midrule
	\multicolumn{1}{l|}{$\eta\,[\%]$}    
	& $\bb{94.5}$ & $25.8$ & $9$ & $\bb{51.6}$ & $14$ & $17.9$ \\
	\multicolumn{1}{l|}{$T_{alg}\,\mathrm{[s]}$} 
	& $\bb{6.5}\pm \bb{4.5}$ & $7.2 \pm 1.6$ & $7.4\pm 3.4$ & $16.5\pm 7.7$ & $\bb{8.9}\pm \bb{1.9}$ & $9.5\pm 4.4$ \\
	\multicolumn{1}{l|}{$D\,\mathrm{[rad]}$}   
	& $16.1\pm 11.8$ & $\bb{13.1}\pm \bb{3.9}$ & $13.5\pm 3.1$ & $37.4\pm 16.8$ & $\bb{14.4}\pm \bb{4.4}$ & $18.4\pm 4.4$ \\ \bottomrule
\end{tabular}
	}
	\label{tab_DRGBT_additional_scenarios_results}
\end{table}

\vspace{-0.3cm}
\subsection{Scalability beyond 6-DoFs}
To investigate the impact of dimensionality on algorithm performance, the competing methods are evaluated on a spatial $n$-DoF manipulator, where $n \in \{10,18\}$ (Scenario 3 in Fig. \ref{fig_DRGBT_additional_scenarios_comparison} depicts spatial 14-DoF robot). The environment is populated with five randomly placed obstacles, each with fixed dimensions of $30 \times 30 \times 30\,\mathrm{[cm^3]}$. Obstacles motion undergoes the same principles as described in Subsec. \ref{Subsec. Scenario Setup}. For each of 1000 trials, the linear velocity of each obstacle is randomly chosen from the range $(0, 0.3]\,\mathrm{[\frac{m}{s}]}$. The maximal allowable runtime per trial is fixed to $30\,\mathrm{[s]}$.

Tab. \ref{tab_DRGBT_high_DoF_scenarios_results_10dof_18dof} shows the obtained results. In parentheses next to success rate, the percentage of runs is depicted when the algorithm runtime exceeded without colliding with obstacles. Such measure reveals that collisions for DRGBT occur in less than $4\, [\%]$ cases. Clearly, the proposed approach displays dominant performance in the case of 10-DoF robot. As for 18-DoF robot, DRGBT achieves a higher success rate than competing methods, but both $\eta$ and $T_{alg}$ turn out to be worse, which indicates a substantial decrease in these performance metrics w.r.t. number of joints $n$. Moreover, the practicality of all the algorithms become questionable since waiting for several minutes for the robot to reach the goal might be unacceptable. Regardless, DRGBT appears to be superior in term of success rate and avoiding collisions within the tested scenarios with higher values of $n$. To better support this claim, clearly a more comprehensive investigation is necessary over a wider range of manipulator types and obstacle counts.

\begin{table}[t]
	\caption{Results for Scenario 3 from the comparison study.}
	\vspace{-0.2cm}
	\centering
	\resizebox{0.495\textwidth}{!}{%
		\begin{tabular}{@{}lccc|ccc@{}}
	\cmidrule(l){2-7}
	& \multicolumn{3}{c|}{10-DoF robot} & \multicolumn{3}{c}{18-DoF robot} \\ 
	\cmidrule(l){2-7} 
	& DRGBT     & MARS & RRT$^\mathrm{X}$  & DRGBT     & MARS  & RRT$^\mathrm{X}$  \\ \midrule
	\multicolumn{1}{l|}{$\eta\,[\%]$}    
	& $\bb{89.1}$ ($\bb{7.3}$) & $68$ ($2.1$) & $44.6$ ($4.3$) & $\bb{41.2}$ ($\bb{57.1}$) & $37.5$ ($45.2$) & $29.4$ ($48$) \\
	\multicolumn{1}{l|}{$T_{alg}\,\mathrm{[s]}$} 
	& $\bb{8.6} \pm \bb{7.6}$ & $12.1\pm 7.4$ & $9.7\pm 4.9$ & $16.9\pm 11.4$ & $20.6\pm 7.7$ & $\bb{10.1\pm 5.3}$ \\
	\multicolumn{1}{l|}{$D\,\mathrm{[rad]}$}   
	& $\bb{20.1}\pm \bb{14.4}$ & $23.0\pm 12.4$ & $21.2\pm 6.1$ & $48.3\pm 29.3$ & $41.9\pm 14.6$ & $\bb{30.2\pm 7.4}$ \\ \bottomrule
\end{tabular}
	}
	\label{tab_DRGBT_high_DoF_scenarios_results_10dof_18dof}
\end{table}

Redundancy constitutes a critical factor in robotics, necessitating careful consideration. Many high-DoF manipulators exhibit kinematic redundancy, which can be exploited to optimize secondary objectives such as manipulability, energy efficiency, joint limit avoidance, and collision avoidance. Integrating redundancy resolution techniques, including null-space projection methods \cite{xing2020enhancement, xing2022dual} or optimization-based approaches \cite{kanoun2011kinematic}, could further extend the applicability of the proposed DRGBT algorithm to redundant systems. The results obtained in relatively high-dimensional $\mathcal{C}$-spaces within Tab. \ref{tab_DRGBT_high_DoF_scenarios_results_10dof_18dof} demonstrate the potential of DRGBT for effectively addressing redundancy. However, the present work exclusively formulates goal configurations within the robot's $\mathcal{C}$-space. The incorporation of methods that define goals in the robot's workspace is left for future investigation.

\vspace{-0.3cm}
\section{Experiments}
\label{Sec. Experiments}
We conduct several experiments to evaluate the real-time performance and effectiveness of DRGBT(-safe) algorithm. First, we describe the used hardware and experimental setup, and then discuss the obtained results.

\begin{figure*}[t]
	\centering
	\includegraphics[width=\linewidth]{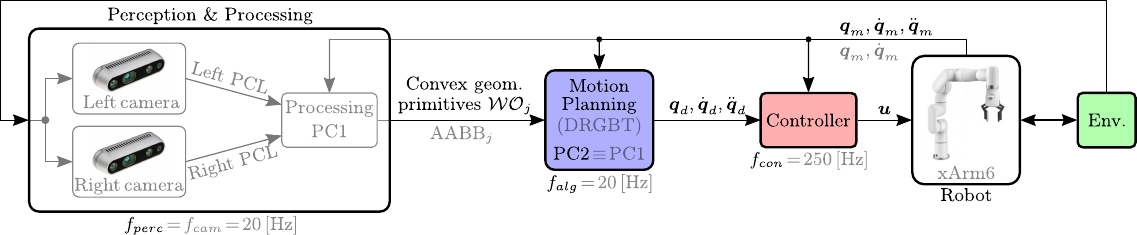}
	\vspace{-0.5cm}
	\caption{Architecture of the used real system. Gray text/lines/blocks indicate specific features used within our experimental study. Black components denote general functionalities supported by the proposed approach.}
	\label{fig_real_setup_scheme}
	\vspace{-0.3cm}
\end{figure*}

\begin{figure*}[t]
	\centering
	\includegraphics[width=\linewidth]{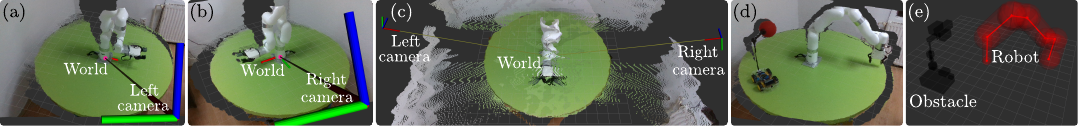}
	\vspace{-0.5cm}
	\caption{The view from both the left (a) and right camera (b), as well as their overlapped/combined view (c). The obstacle -- a car with the balloon (d) is rendered to 14 AABBs (e). Bounding capsules and a wired model of the manipulator (e).}
	\label{fig_real_setup}
	\vspace{-0.6cm}
\end{figure*}

\vspace{-0.3cm}
\subsection{Experimental Setup and Implementation Details}
\label{Subsec. Experimental Setup and Implementation Details}
Fig. \ref{fig_real_setup_scheme} illustrates the architecture of the used real system. We use two \textit{Intel RealSense D435i} depth cameras to perceive the environment with the frequency $f_{perc} = f_{cam} = 20\,\mathrm{[Hz]}$. Each of them provides a \textit{point cloud}, which are then combined and processed. Fig. \ref{fig_real_setup} reveals the angle of view from both the left (a) and the right camera (b), as well as their overlapped/combined view (c). The ``Processing'' block exploits Point Cloud Library (PCL) \cite{rusu20113d}, which combines both left and right point clouds into a single unified point cloud, and then performs the following steps: downsampling, filtering, removing outliers, removing the robot from the scene, clustering and segmentation. The downsampling leads to a voxel-grid-based representation with the unit cell being a $2\,\mathrm{[cm]}$ side cube. The table points are removed based on a priori knowledge that it is a static obstacle. The robot points are also filtered out by reading the currently measured joint configuration $\bb{q}_m$ from proprioceptive sensors. 
The remaining point cloud points are then grouped into clusters, and the segmentation process is performed. As a result, each cluster is transformed into a single axis-aligned bounding box (AABB). The actual position and size of each AABB are adapted using a specified tolerance in a way that such resulting AABB encapsulates all points from the cluster. It is worth noting that the proposed architecture does not necessarily require the representation of obstacles via AABBs. In general, the obstacles can be represented by any convex geometric primitives. In this context, AABBs are used for mere simplicity. The implementation of ``Processing'' block is available \href{https://github.com/robotics-ETF/xarm6-etf-lab/blob/main/src/etf_modules/perception_etflab}{here}, with a simple example shown in Fig. \ref{fig_real_setup} (d), where the obstacle (a car with the balloon) is rendered to 14 clusters and therefore 14 AABBs (black boxes in (e)). 

For experiments, we use the UFactory xArm6 manipulator. 
As mentioned in Subsec. \ref{Subsec. Scenario Setup}, the robot's links are approximated by bounding capsules, which are depicted in red by Fig. \ref{fig_real_setup} (e) for the assumed configuration (d). Moreover, a wired model of the robot is indicated by the red line segments. A low-level controller ``Controller'' provides a control signal $\bb{u}$ to the robot, and requires desired and measured values for position $\bb{q}_d$ and $\bb{q}_m$, velocity $\dot{\bb{q}}_d$ and $\dot{\bb{q}}_m$ (and possibly acceleration $\ddot{\bb{q}}_d$ and $\ddot{\bb{q}}_m$), respectively, of each robot's joint with the controlled frequency $f_{con} = 250\,\mathrm{[Hz]}$. Specifically, UFactory xArm6 only provides $\bb{q}_m$ and $\dot{\bb{q}}_m$.

The planning algorithm is implemented within the ``Planning'' block\footnote{The algorithm is run on the laptop PC with Intel\textregistered\, Core\texttrademark\, i7-9750H CPU @ 2.60 GHz $\times$ 12 with 16 GB of RAM, with the code compiled to run on a single core of the CPU without any GPU acceleration. The affordability of systems such as Intel NUC PCs, commonly found in robotics applications, give us the idea of what can be considered as a low-cost x86 CPU. In experiments, we use the same PC within the ``Processing'' block, i.e., $\mathrm{PC1 \equiv PC2}$. Generally, they can be different.} using ROS2 environment with the open-source code available online \href{https://github.com/robotics-ETF/xarm6-etf-lab}{here}. Besides standard inputs required by the planning, its output is a desired trajectory containing values for $\bb{q}_d$, $\dot{\bb{q}}_d$ and $\ddot{\bb{q}}_d$ of each robot's joint. Although the planning generates trajectories with a frequency $f_{alg}$, which is usually lower than $f_{con}$, the obtained trajectory can be sampled with $f_{con}$, and then fed to the controller with the corresponding samples in order to achieve a smooth motion of the robot. Since $f_{cam}$ may be lower than $f_{alg}$, it is questionable whether setting $f_{alg} > f_{cam}$ makes sense in practical applications. Theoretically, DRGBT can execute in such cases, since the last available camera measurements can always be used. However, it turns out that using $f_{alg} \gg f_{cam}$ (e.g., $f_{alg} \geq 3f_{cam}$) is not recommended since the algorithm may falsely perceive the obstacles as stationary for a few iterations, thus causing generated plans to be inadequate. This might be addressed by predicting the motion of obstacles (e.g., \cite{ragaglia2018trajectory, hadzic2023}), however, this exceeds the scope of this paper. Since the simulation study suggests the sweet spot to be around $f_{alg}^* = 20\,\mathrm{[Hz]}$, the cameras' frame rate is also set to $f_{cam} = 20\,\mathrm{[Hz]}$ for all conducted scenarios in the sequel. Another reason for setting such a value of $f_{cam}$ is achieving a hard real-time execution of the ``Processing'' block w.r.t. specified voxel grid resolution.

\subsection{Obtained Results}
\label{Subsec. Obtained Results}
\begin{figure*}[t]
	\centering
	\includegraphics[width=\linewidth]{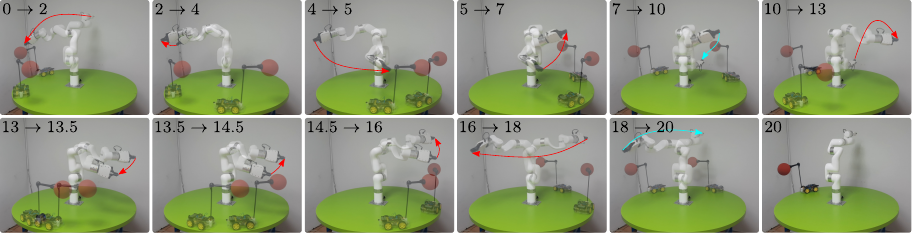}
	\caption{Snapshots from Scenario 1 -- ``Moving-car with balloon''.}
	\label{fig_real_scenario_car}
	\vspace{-0.45cm}
\end{figure*}

\begin{figure*}[t]
	\centering
	\includegraphics[width=\linewidth]{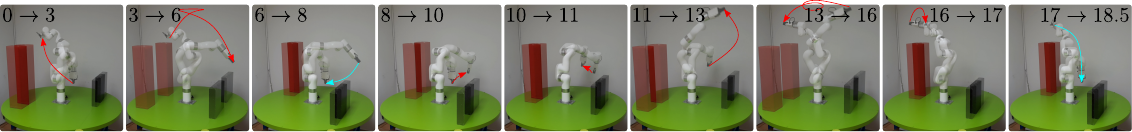}
	\caption{Snapshots from Scenario 2 -- ``Moving-boxes''.}
	\label{fig_real_scenario_boxes}
	\vspace{-0.4cm}
\end{figure*}

\begin{figure*}[t]
	\centering
	\includegraphics[width=\linewidth]{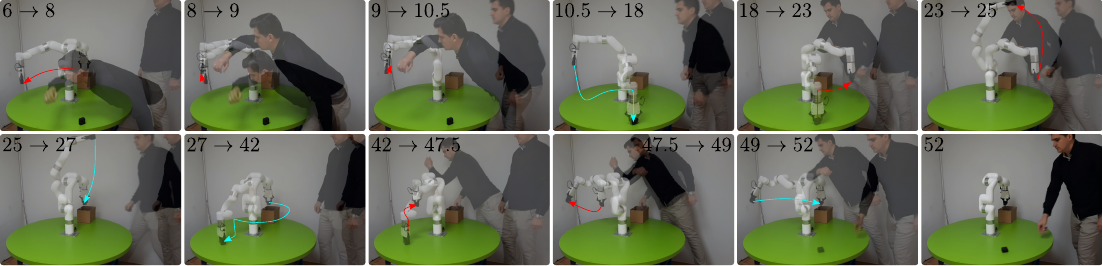}
	\caption{Snapshots from Scenario 3 -- ``Human as an obstacle''.}
	\label{fig_real_scenario_human}
	\vspace{-0.4cm}
\end{figure*}

\begin{figure*}[t]
	\centering
	\includegraphics[width=\linewidth]{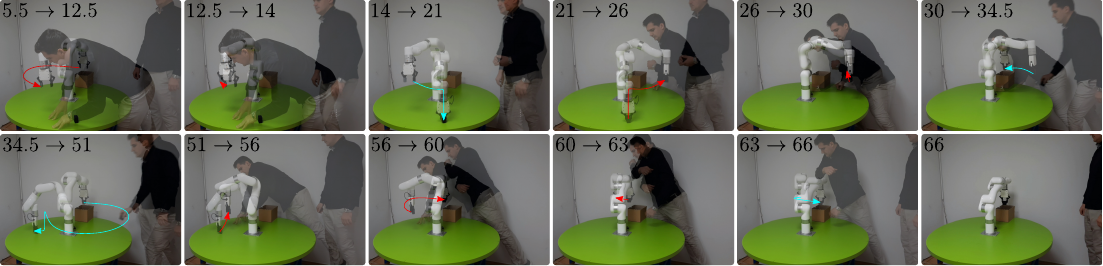}
	\caption{Snapshots from Scenario 4 -- ``Safe collaboration/coexistence with human''.}
	\label{fig_real_scenario_human_safe}
	\vspace{-0.7cm}
\end{figure*}

To validate the proposed motion planning method in dynamic environments, we conduct four real scenarios (video available \href{https://www.youtube.com/watch?v=lG7q0PuFhG0}{here}). Representative snapshots are shown in Figs. \ref{fig_real_scenario_car}, \ref{fig_real_scenario_boxes}, \ref{fig_real_scenario_human} and \ref{fig_real_scenario_human_safe}, where the path traversed by the end-effector is depicted by red lines. The exception is the cyan line that denotes the part of the trajectory before the goal is reached. Time instances $t$ in $\mathrm{[s]}$ are marked in the top part of each figure. The first three scenarios test the regular variant of DRGBT, while the last one deals with DRGBT-safe (from Sec. \ref{Sec. Safe Motion of the Robot in Dynamic Environments Under Bounded Obstacle Velocity}).

\begin{figure}[t]
	\centering
	\includegraphics[width=\linewidth]{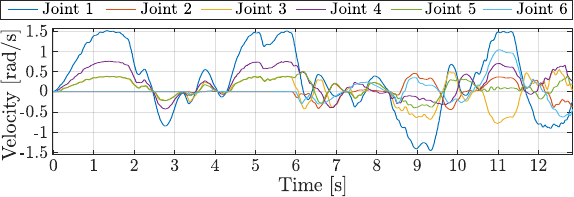}
	\vspace{-0.55cm}
	\caption{Measured velocities for each joint from Scenario 1.}
	\label{fig_velocities_measured}
\end{figure}

Scenario 1 -- ``Moving-car with balloon'' is shown in Fig. \ref{fig_real_scenario_car}, and considers a car with the balloon circulating counterclockwise around the robot with a constant speed of cca. $0.36\,\mathrm{[\frac{m}{s}]}$ (i.e., the circle with a radius of $58\,\mathrm{[cm]}$ is completed within $10\,\mathrm{[s]}$). For the start and goal we choose $\bb{q}_{start} = \left[-\pi, 0, -\frac{\pi}{2}, 0, 0, 0\right]^T$ and $\bb{q}_{goal} = \left[\pi, 0, 0, \pi, \frac{\pi}{2}, 0\right]^T$, respectively, and they are swapped when the goal is reached thus emulating repetitive pick-and-place operations. The robot follows an initially planned path during the time $t\in[0,2]$, and then it decelerates in order not to hit the obstacle, after which the replanning is triggered. Similar situation occurs for $t\in[4,5]$ and $t\in[5,7]$. After $10\,\mathrm{[s]}$, the goal is reached. During $t \in [10, 13.5]$, the robot and obstacle approach each other. However, the robot successfully avoids the obstacle within $t \in [13.5, 16]$. Afterward, the path is replanned, and the goal is therefore reached since the predefined path remains collision-free. The measured joint velocities for Scenario 1 are given in Fig. \ref{fig_velocities_measured}. Clearly, they remain within the set limits of $1.5\,\mathrm{[\frac{rad}{s}]}$ for each joint. Analogous diagrams are obtained for other scenarios, yet those are omitted for the limited space. 

Fig. \ref{fig_real_scenario_boxes} shows Scenario 2 -- ``Moving-boxes'', which considers two boxes moving alternately with a constant speed of cca. $0.2\,\mathrm{[\frac{m}{s}]}$. For the start and goal, we choose $\bb{q}_{start} = \left[-\pi, 0, -\frac{\pi}{4}, 0, \frac{\pi}{4}, 0\right]^T$ and $\bb{q}_{goal} = \left[\pi, 0, -\frac{\pi}{4}, 0, \frac{\pi}{4}, 0\right]^T$, respectively, which are swapped when the goal is reached. The robot successfully follows an initially planned path, yet replanning has been triggered in cca. $t\in\{1,3,6\}$. After reaching the goal in $t = 8$, the black obstacle is efficiently avoided during $t\in[10,11]$. 
Thereafter, a new predefined path is followed for $t\in[11,16]$, after which the robot bypasses the red obstacle multiple times while taking replanning actions which lead the robot quickly to the goal.

Scenario 3 -- ``Human as an obstacle'' from Fig. \ref{fig_real_scenario_human} and Scenario 4 -- ``Safe collaboration/coexistence with human'' from Fig. \ref{fig_real_scenario_human_safe} represent a mock-up of scenario where the robot performs pick-and-place operations while the human operator interrupts the robot by frequent intrusions into the workspace. 
The robot has to pick up objects from the table while avoiding the human at the same time. To facilitate target object recognition, the robot is programmed to pick only relatively small objects while those larger ones are considered dynamic obstacles. The maximally expected obstacle velocity in Scenario 4 is set to $v_{obs} = 0.5\,\mathrm{[\frac{m}{s}]}$, which is particularly relevant to generating $\mathcal{DEB}$s, while conducted scenarios with $v_{obs} = 1.6\,\mathrm{[\frac{m}{s}]}$ can be seen in the accompanying \href{https://www.youtube.com/watch?v=lG7q0PuFhG0}{video}.

Generally, when obstacles move more swiftly as in Scenarios 3 and 4, the replanning process is triggered more frequently. As for DRGBT regular variant, the robot always avoids the human (e.g., for $t\in[8, 10.5]$, $t\in[23, 25]$, and $t\in[47.5, 49]$ in Fig. \ref{fig_real_scenario_human}), and then tries to reach the goal configuration when no risk of collision is observed within a certain range (e.g., for $t\in[10.5, 18]$, $t\in[25, 27]$, $t\in[27, 42]$, and $t\in[49, 52]$ in Fig. \ref{fig_real_scenario_human}). On the other hand, DRGBT-safe prioritizes the deceleration of the robot in order not to collide with the human. In case a collision occurs, it will be at zero speed of the robot (type II collision). Interestingly, if the distance-obstacles proximity remains relatively small (or zero), the robot will not be able to move fast (or to move at all) ensuring human safety (e.g., for $t\in[12.5, 14]$, $t\in[26, 30]$, and $t\in[60, 66]$ in Fig. \ref{fig_real_scenario_human_safe}). The robot remains considerably slow until the distance to obstacles increases. It is worth mentioning the robot could not stop timely in $t = 60$ since the estimated human arm velocity reached nearly $1.8\,\mathrm{[\frac{m}{s}]}$, which is substantially higher than the assumed limit of $v_{obs} = 0.5\,\mathrm{[\frac{m}{s}]}$. Nevertheless, \href{https://www.youtube.com/watch?v=lG7q0PuFhG0}{video} reveals that increasing $v_{obs}$ up to $1.6\,\mathrm{[\frac{m}{s}]}$ does not yield collisions. Indeed, both variants of DRGBT successfully complete the required pick-and-place task clearly revealing the trade-off between performance and safety.

\vspace{-0.25cm}
\section{Discussion and Conclusions}
\label{Sec. Discussion and Conclusions}
\vspace{-0.05cm}
The primary limitations of the proposed approach stem from its reliance on the generalized bur of free $\mathcal{C}$-space \cite{lacevic2020gbur}. This concept is mainly applicable to open-chain manipulators and cannot be directly extended to other types of robots. 
However, developing (dynamic) generalized burs applicable to a broader range of robots remains an open research challenge. Another drawback is the implicit assumption that the environment consists of a finite set of convex obstacles. 

Furthermore, Theorem \ref{Theorem Guaranteed safe motion of the robot in dynamic environments under bounded obstacle velocity} requires setting the upper bound on obstacle velocity in order to guarantee safe motion of the robot in dynamic environments. In addition, new obstacles are not allowed to suddenly appear anywhere in the workspace, with the exception of the region $\mathcal{W}_{occ}$ (see Sec. \ref{Sec. Safe Motion of the Robot in Dynamic Environments Under Bounded Obstacle Velocity}).

Finally, the time parameterization of real-time scheduling depends on the specific hardware setup. In case the approach needs to be implemented on another platform and tuned for optimal success rate, a dedicated simulation study is recommended for establishing the corresponding sweet spots.

To summarize, this paper introduces the upgraded version of DRGBT algorithm emphasizing its hard real-time execution capabilities. Accordingly, the scheduling framework is examined through two main tasks, alongside its constitutive routines. Theoretical computation, supported by the randomized trial of up to 50 obstacles, reveals that the algorithm is capable of real-time operation at frequencies up to $100\,\mathrm{[Hz]}$. Moreover, DRGBT outperforms the competing algorithm according to both the success rate and the algorithm execution time.

Furthermore, new structures -- dynamic expanded bubble and dynamic (generalized) bur are formulated and exploited to impose sufficient conditions for a guaranteed safe motion of the robot under certain kinematic constraints. The experiments are carried out through different scenarios including human-robot collaboration. Results confirm that the proposed approach can deal with unpredictable dynamic obstacles in real-time, while simultaneously ensuring safety to the environment.

Future work directions include explicit handling of non-convex obstacles. It would be interesting to investigate the influence of enhanced perception through the use of dedicated hardware (e.g., a separate PC for processing). Finally, tighter integration with the speed and separation monitoring paradigm is desirable for human-robot coexistence contexts.

\vspace{-0.25cm}
\section*{Acknowledgment}
\vspace{-0.05cm}
This work has been supported by Ministry of education and science of Canton Sarajevo, grant no. 11-05-14-27164-1/19.

\vspace{-0.25cm}
\bibliographystyle{IEEEtran}
\bibliography{../../Literatura/References} 


\vspace{-1.2cm}
\begin{IEEEbiography}[{\includegraphics[width=1in,height=1.25in,clip,keepaspectratio]{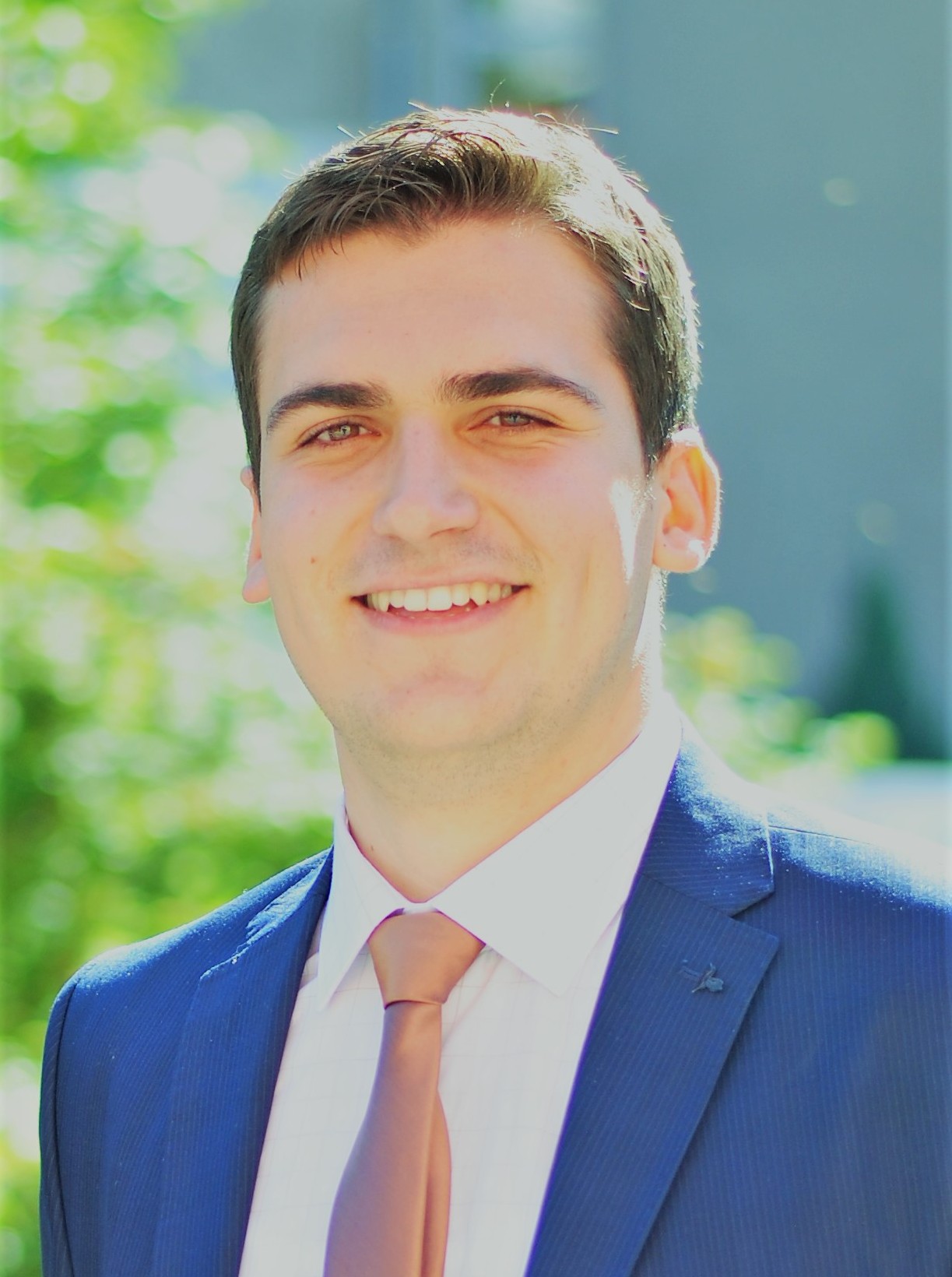}}]
	{Nermin Covic} (Student Member, IEEE) received the B.S. and M.S. degrees, both with highest honors, at the Department for Control Systems and Electronics within the Faculty of Electrical Engineering from University of Sarajevo in 2016 and 2018, respectively. He is currently working as a research and teaching assistant at the same institution, where he is pursuing his Ph.D. degree. His research interests include robotic motion planning, optimization, automation, control systems and system identification.
\end{IEEEbiography}
\vspace{-1.5cm}

\begin{IEEEbiography}[{\includegraphics[width=1in,height=1.25in,clip,keepaspectratio]{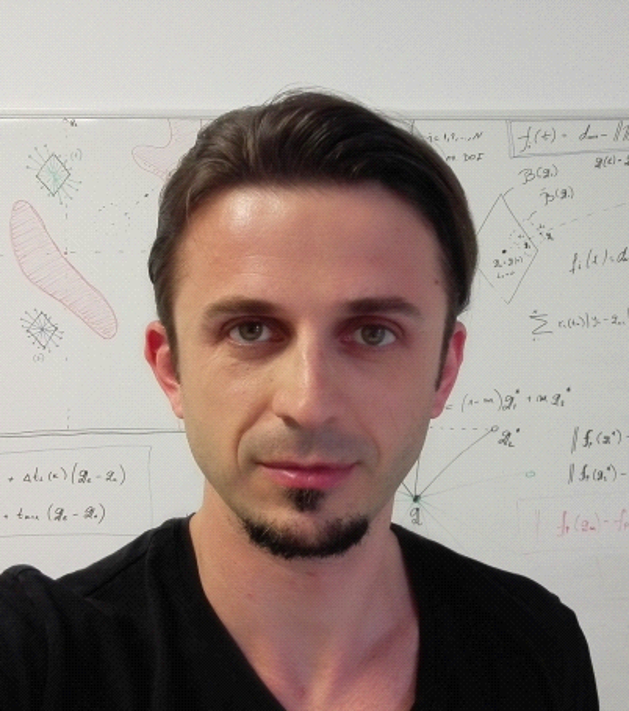}}]
	{Bakir Lacevic} (Member, IEEE) received the Dipl.-Ing. and Magister degrees in automatic control from the University of Sarajevo, Sarajevo, Bosnia and Herzegovina, in 2003 and 2007, respectively, and the Ph.D. degree in information technology from the Politecnico di Milano, Milano, Italy, in 2011. Currently, he is a Professor with the Faculty of Electrical Engineering, University of Sarajevo. His research interests include robotic motion planning, human–robot collaboration, optimization, and machine learning. 	
\end{IEEEbiography}
\vspace{-1.2cm}

\begin{IEEEbiography}[{\includegraphics[width=1in,height=1.25in,clip,keepaspectratio]{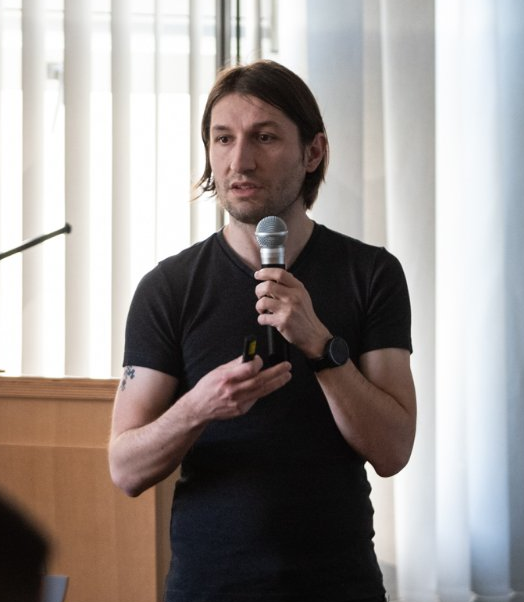}}]
	{Dinko Osmankovic} (Member, IEEE) was an Associate Professor at the Faculty of Electrical Engineering, University of Sarajevo. He obtained his PhD at the same institution in 2015 for the work in the field of Automated 3D thermal modeling of indoor environments. Since then, he’s been working in the fields of Computer Vision, Motion Planning and Artificial Intelligence in various applications. Currently he works as a Software Engineer for Agile Robots SE.
\end{IEEEbiography}
\vspace{-1.5cm}

\begin{IEEEbiography}[{\includegraphics[width=1in,height=1.25in,clip,keepaspectratio]{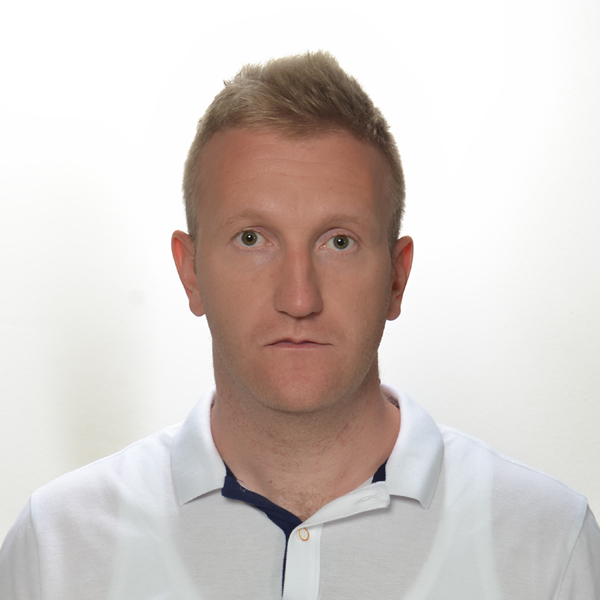}}]
	{Tarik Uzunovic} (Senior Member, IEEE) received the B.Eng. and M.Eng. degrees in electrical engineering from the University of Sarajevo, Sarajevo, Bosnia and Herzegovina, in 2008 and 2010, respectively, and the Ph.D. degree in mechatronics from Sabanci University, Istanbul, Turkey, in 2015.
	
	He is currently a Full Professor with the Department of Automatic Control and Electronics, Faculty of Electrical Engineering, University of Sarajevo. His research interests include control theory, motion control, robotics, and mechatronics.
\end{IEEEbiography}

\end{document}